\documentclass[11pt,a4paper]{article}
\usepackage[T1]{fontenc}
\usepackage[utf8]{inputenc}
\usepackage{lmodern}
\usepackage{geometry}
\usepackage{graphicx}
\usepackage{booktabs}
\usepackage{multirow}
\usepackage{bm}
\usepackage{float}
\usepackage{tabularx}
\usepackage{array}
\usepackage{algorithm}
\usepackage{algpseudocode}
\usepackage{amsmath,amssymb}
\usepackage[authoryear,round]{natbib}
\usepackage[hidelinks]{hyperref}
\newcolumntype{P}[1]{>{\raggedright\arraybackslash}p{#1}}
\newcolumntype{Y}{>{\raggedright\arraybackslash}X}
\title{RUL-Aware RRT*: Degradation-Balanced Motion Planning for Robotic Manipulators}
\author{Haibo Li$^{1}$ \quad Zhiguo Zeng$^{2}$\thanks{Corresponding author: zhiguo.zeng@centralesupelec.fr} \quad Xu Li$^{3}$\\[0.5em]
\small $^{1}$Laboratoire G\'enie Industriel, CentraleSup\'elec, Universit\'e Paris-Saclay, France\\
\small $^{2}$Chair of Risk and Resilience of Complex Systems, CentraleSup\'elec, Universit\'e Paris-Saclay, France\\
\small $^{3}$Key Laboratory of Urban Underground Engineering, Beijing Jiaotong University, China}
\date{}
\begin{document}
\maketitle

\begin{abstract}
Robotic manipulators operating over long durations often experience uneven joint degradation, which causes the weakest actuator to fail prematurely, leads to unplanned downtime, and results in significant operational losses. Traditional motion-planning algorithms do not account for joint health conditions and therefore tend to exacerbate this imbalance during extended operation. To address this challenge, this study introduces the RUL-aware RRT*, a motion-planning method that incorporates joint remaining useful life information into the planning process and adaptively adjusts joint usage in response to evolving health conditions. The method is evaluated across three representative scenarios, namely the Full Health Scenario (FHS), the Heterogeneous Degradation Scenario (HDS), and the Local Degradation Scenario (LDS). The results show that the RUL-aware RRT* effectively suppresses degradation imbalance, delays the emergence of bottleneck failures, and improves the long-term reliability of the robotic system during extended operation. These findings demonstrate that integrating health feedback into motion planning provides a practical and robust pathway for enhancing the durability and operational resilience of manipulators subject to continuous wear.

\end{abstract}


\section{Introduction}\label{sec:1}

In modern manufacturing, industrial robots have become essential equipment for achieving automated production and flexible manufacturing\citep{russmann2015industry}. As the most critical component of industrial robots, multi-degree-of-freedom robotic arms rely on multiple joint motors that must operate in close coordination to perform complex, continuous, and high-precision tasks \citep{fan2019autonomous}. However, in practical operation, motion planning is typically optimized for geometric efficiency or energy performance \citep{noreen2016optimal, feng2024adaptive}, while differences in joint health conditions and the resulting uneven motion usage across actuators are often overlooked. This leads to uneven motor utilization \citep{raviola2021harmonic}, where certain joints may experience accelerated degradation due to high workloads or frequent movements, eventually becoming the weakest link of the system \citep{kot2021method}. Once such a motor fails, the entire robot must be shut down, resulting in unplanned downtime and seriously affecting production continuity and equipment reliability \citep{lei2025condition, mahyari2018domain, raviola2021harmonic}. This issue, in which local degradation triggers overall system failure, has become one of the key challenges in the reliability management of industrial robots.  

To address such issues, extensive research has been conducted in recent years in both academia and industry within the fields of condition monitoring and predictive maintenance (PdM) based on robotic arms \citep{lei2025condition, moore2024equipment, uhlmann2020condition}. By leveraging remaining useful life (RUL) prediction and digital twin technologies, researchers are able to monitor equipment health in real time, predict the degradation trends of critical components, and make informed maintenance decisions accordingly \citep{mourtzis2023robotic, borgi2017data}. These approaches have significantly enhanced the intelligence of maintenance management, shifting it from a reactive mode to a proactive strategy \citep{molkeda2023corrective}. However, most existing studies remain at the monitoring and scheduling layers, where the focus lies in sensing system health and determining maintenance timing \citep{kumar2023prognostics}. Their outputs, such as health indicators, degradation trajectories, or maintenance schedules, are seldom integrated into the motion-planning layer, preventing health information from directly influencing path generation or motion decisions \citep{schacht2018path}. In other words, although health information is effectively perceived, it is rarely incorporated into the robot’s path generation and motion decision-making processes, thereby limiting the realization of a closed-loop predictive maintenance framework.

On the other hand, sampling-based motion planning algorithms, such as RRT, RRT*, and PRM, provide a flexible framework for incorporating multidimensional constraints into the planning process \citep{kingston2018sampling, zhang2025motion, luo2023survey}. In recent years, various studies have introduced factors such as energy and consumption \citep{alam2023novel}, collision risk \citep{liu2023radius}, safety distance \citep{mainprice2011planning}, battery endurance \citep{chitre2023enhanced} into the cost functions of these algorithms, thereby achieving a comprehensive optimization of efficiency, safety, and reliability. These works demonstrate that path planning can be applied not only for geometric feasibility analysis but also holds great potential for integrating health-related factors. However, existing research has not yet systematically incorporated the uneven degradation conditions across multiple motors and their imbalanced RUL distribution into planning models, leaving robotic systems lacking health awareness at the path-planning level.

Based on the above analysis, three major research gaps can be identified. First, traditional path-planning algorithms often lead to uneven motor degradation, as they optimize only geometric or efficiency-related objectives without considering motor health conditions. Second, predictive maintenance techniques largely remain at the monitoring and scheduling levels, without direct integration into the motion-planning process. Third, existing path-planning studies mainly focus on energy efficiency or collision avoidance, while overlooking the uneven degradation conditions across multiple motors and the resulting lifetime bottleneck.

To address these challenges, this study proposes an RUL-aware RRT* motion-planning algorithm that aims to redistribute joint-level motion usage, enhance health awareness, and mitigate lifetime bottlenecks. The proposed algorithm is developed on the basis of the traditional RRT* framework \citep{karaman2011sampling}, whose original cost function considers only the path length, while RUL-aware RRT* extends it by introducing an RUL-aware cost function that incorporates joint health information into the path-cost formulation, thereby enabling deep integration of health awareness into the motion-planning process. In addition, two key mechanisms are designed within the framework: (1) an Online RUL Update Mechanism,which dynamically updates the current remaining useful life (RUL) of each joint before each motion-planning process; and (2) an RUL-driven Adaptive Joint Weighting Mechanism, which adaptively balances motor loads under different degradation conditions to achieve health-aware motion planning.

In multi-scenario experiments conducted on a UR5 six-degree-of-freedom robotic arm, including the Full Health Scenario, the Heterogeneous Degradation Scenario, and the Local Degradation Scenario, the results demonstrate that the proposed method achieves balanced degradation, enhances health-awareness, and extends the first-replacement time of critical motors, thereby realizing planable predictive maintenance. This study provides a new perspective for achieving health awareness and lifetime balancing at the task-execution level of industrial robots, and lays the foundation for future research on reliability-aware motion planning.

The implementation and experimental data are available at \url{https://github.com/HelpLee/MotionPlanning_RULRRTstar}.





























%
%
\section{Background and Preliminaries}\label{sec:2}

Sampling-based motion planning algorithms have been widely adopted in robotic systems
due to their ability to handle high-dimensional configuration spaces
without requiring explicit geometric representations of obstacles.
Among these methods, the Rapidly-exploring Random Tree (RRT) algorithm 
and its optimal variant RRT* have become the most prominent frameworks 
for generating feasible and near-optimal trajectories efficiently. 
RRT* is particularly suitable for robotic manipulators, 
as it incrementally constructs a connected tree in the configuration space 
while refining the solution toward the global optimum through local rewiring. 
Its asymptotic optimality and probabilistic completeness make it 
a robust foundation for motion planning under complex kinematic constraints.

\begin{figure}
    \centering
    \includegraphics[width=0.7\linewidth]{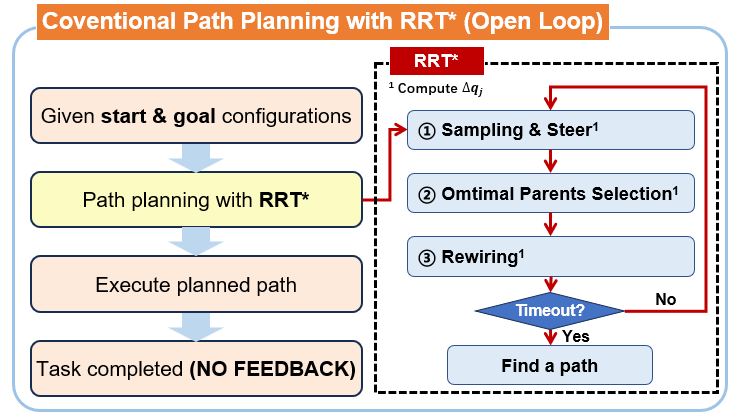}
    \caption{Conventional path planning with RRT* (open-loop framework)}
    \label{fig:rrtstar}
\end{figure}

We build upon the standard RRT* algorithm, a sampling-based motion-planning method that incrementally constructs a connected tree 
\( G = (V, E) \) in the configuration space \( \mathcal{X} \subset \mathbb{R}^J \). 
The planning problem is formulated as finding a collision-free path 
within the free configuration space \( \mathcal{X}_{\mathrm{free}} \subset \mathcal{X} \) 
that connects a given start configuration \( x_{\mathrm{init}} \in \mathcal{X}_{\mathrm{free}} \) 
to a goal region \( \mathcal{X}_{\mathrm{goal}} \subset \mathcal{X}_{\mathrm{free}} \).
Starting from \( x_{\mathrm{init}} \), RRT* repeatedly samples random states, 
extends the tree toward these samples, 
and rewires the connections to minimize the cumulative path cost 
while ensuring that the tree eventually reaches \( \mathcal{X}_{\mathrm{goal}} \). 
Through this iterative process, the algorithm converges asymptotically 
to an optimal solution while maintaining probabilistic completeness.
The overall procedure is summarized in Algorithm~\ref{alg:RRTstar}, 
and more implementation details can be found in the original work 
by Karaman and Frazzoli~\citep{karaman2011sampling}. 
In each iteration, the planner performs three main steps:

\medskip
· \textbf{Expansion (Lines 3--6).}  
A random sample \(x_{\mathrm{rand}}\) is drawn from the free configuration space. 
The algorithm then identifies the nearest node \(x_{\mathrm{nearest}}\) in the existing tree by evaluating the path cost between each vertex and the sampled configuration. 
Formally, the nearest node is determined as
\begin{equation}
    x_{\mathrm{nearest}} = \arg\min_{x_i \in V} 
    c(\text{Line}(x_i, x_{\mathrm{rand}})),
    \label{eq:rrt_nearest_cost}
\end{equation}
where \(c(\cdot)\) represents the cost of connecting two configurations. 
Once \(x_{\mathrm{nearest}}\) is identified, a new node \(x_{\mathrm{new}}\) is generated by steering from \(x_{\mathrm{nearest}}\) toward \(x_{\mathrm{rand}}\). 

\medskip
· \textbf{Parent Selection (Lines 7--13).}  
If the edge $(x_{\mathrm{nearest}}, x_{\mathrm{new}})$ is collision-free, 
a set of neighboring nodes $X_{\mathrm{near}}$ around $x_{\mathrm{new}}$ 
is identified according to a predefined connection radius. 
The parent node $x_{\min}$ is then selected as the one that minimizes 
the total path cost from the start configuration to the new node:
\begin{equation}
x_{\min} =
\arg\min_{x_{\mathrm{near}} \in X_{\mathrm{near}}}
\big[\mathrm{Cost}(x_{\mathrm{near}}) 
+ c(\mathrm{Line}(x_{\mathrm{near}}, x_{\mathrm{new}}))\big],
\label{eq:rrt_parent_selection}
\end{equation}
Here, $\mathrm{Cost}(x)$ denotes the accumulated cost of the unique tree path 
from the start configuration $x_{\mathrm{init}}$ to node $x$, 
computed as the sum of the incremental costs along that path:
\begin{equation}
\mathrm{Cost}(x)
= \sum_{(x_i, x_{i+1}) \in \mathcal{P}(x)}
c(\mathrm{Line}(x_i, x_{i+1})),
\label{eq:rrt_accumulated_cost}
\end{equation}
where $\mathcal{P}(x)$ is the ordered sequence of vertices connecting 
$x_{\mathrm{init}}$ and $x$ in the current tree. 
The term $c(\mathrm{Line}(x_{\mathrm{near}}, x_{\mathrm{new}}))$ 
represents the local cost of connecting the neighbor $x_{\mathrm{near}}$ 
to the newly generated node $x_{\mathrm{new}}$, 
typically defined as the Euclidean distance in the configuration space. 
Intuitively, this step selects as the parent the neighbor that provides 
the lowest total cost from $x_{\mathrm{init}}$ to $x_{\mathrm{new}}$ 
through the existing tree.

\medskip
· \textbf{Rewiring (Lines 14--22).}  
After adding $x_{\mathrm{new}}$ to the tree, each node $x_{\mathrm{near}}$ in $X_{\mathrm{near}}$ 
is checked to determine whether its total cost can be reduced by reconnecting through $x_{\mathrm{new}}$. 
If the following condition holds:
\begin{equation}
\mathrm{Cost}(x_{\mathrm{new}}) 
+ c(\mathrm{Line}(x_{\mathrm{new}}, x_{\mathrm{near}})) 
< \mathrm{Cost}(x_{\mathrm{near}}),
\label{eq:rrt_rewire_condition}
\end{equation}
then the edge is rewired accordingly so that $x_{\mathrm{near}}$ becomes a child of $x_{\mathrm{new}}$. 
This local rewiring operation allows the planner to incrementally refine the tree 
and maintain asymptotic optimality as the number of samples increases.

In the standard RRT* framework, all cost terms $c(\cdot)$ and $\mathrm{Cost}(\cdot)$ 
are purely defined by the path length between two configurations. 
For robotic arm manipulators, each configuration is represented by a vector of joint angles
\begin{equation}
\textbf{q} = [q_1, q_2, \dots, q_j]^\top \in \mathbb{R}^J,
\label{eq:joint_vector}
\end{equation}
where $J$ is the number of joints. 
Accordingly, the path-length cost between two configurations $x_i$ and $x_{i+1}$ 
is computed as the Euclidean distance in the joint space:
\begin{equation}
c(\mathrm{Line}(x_i, x_{i+1}))
= \| q_{x_i} - q_{x_{i+1}} \|_2
= \sqrt{\sum_{j=1}^{J} (q_{x_i,j} - q_{x_{i+1},j})^2}.
\label{eq:joint_space_distance}
\end{equation}
Here, $q_{x_1}$ and $q_{x_2}$ denote the joint-angle vectors associated with configurations 
$x_1$ and $x_2$, respectively. 
This Euclidean metric provides a geometric measure of displacement in the manipulator’s 
joint space and is used to evaluate both the incremental edge cost $c(\mathrm{Line}(\cdot))$ 
and the accumulated path cost $\mathrm{Cost}(\cdot)$.

\begin{algorithm}[H]
\caption{RRT*}
\label{alg:RRTstar}
\begin{algorithmic}[1]
\State $V \leftarrow \{x_{\mathrm{init}}\}; \quad E \leftarrow \emptyset$
\For{$i = 1, \ldots, n$}
    \State $x_{\mathrm{rand}} \leftarrow$ \textsc{SampleFree}()
    \State $x_{\mathrm{nearest}} \leftarrow$ \textsc{Nearest}$(G = (V,E), x_{\mathrm{rand}})$
    \State $x_{\mathrm{new}} \leftarrow$ \textsc{Steer}$(x_{\mathrm{nearest}}, x_{\mathrm{rand}})$
    \If{\textsc{ObstacleFree}$(x_{\mathrm{nearest}}, x_{\mathrm{new}})$}
        \State $X_{\mathrm{near}} \leftarrow$ \textsc{Near}$(G = (V,E), x_{\mathrm{new}}, \min\{\gamma_{RRT^*}(\log(\text{card}(V))/\text{card}(V))^{1/d}, \eta\})$
        \State $V \leftarrow V \cup \{x_{\mathrm{new}}\}$
        \State $x_{\mathrm{min}} \leftarrow x_{\mathrm{nearest}}; \quad c_{\mathrm{min}} \leftarrow \text{Cost}(x_{\mathrm{nearest}}) + c(\text{Line}(x_{\mathrm{nearest}}, x_{\mathrm{new}}))$
        \ForAll{$x_{\mathrm{near}} \in X_{\mathrm{near}}$} \Comment{Connect along a minimum-cost path}
            \If{\textsc{CollisionFree}$(x_{\mathrm{near}}, x_{\mathrm{new}}) \land \text{Cost}(x_{\mathrm{near}}) + c(\text{Line}(x_{\mathrm{near}}, x_{\mathrm{new}})) < c_{\mathrm{min}}$}
                \State $x_{\mathrm{min}} \leftarrow x_{\mathrm{near}}$
                \State $c_{\mathrm{min}} \leftarrow \text{Cost}(x_{\mathrm{near}}) + c(\text{Line}(x_{\mathrm{near}}, x_{\mathrm{new}}))$
            \EndIf
        \EndFor
        \State $E \leftarrow E \cup \{(x_{\mathrm{min}}, x_{\mathrm{new}})\}$
        \ForAll{$x_{\mathrm{near}} \in X_{\mathrm{near}}$} \Comment{Rewire the tree}
            \If{\textsc{CollisionFree}$(x_{\mathrm{new}}, x_{\mathrm{near}}) \land \text{Cost}(x_{\mathrm{new}}) + c(\text{Line}(x_{\mathrm{new}}, x_{\mathrm{near}})) < \text{Cost}(x_{\mathrm{near}})$}
                \State $x_{\mathrm{parent}} \leftarrow \text{Parent}(x_{\mathrm{near}})$
                \State $E \leftarrow (E \setminus \{(x_{\mathrm{parent}}, x_{\mathrm{near}})\}) \cup \{(x_{\mathrm{new}}, x_{\mathrm{near}})\}$
            \EndIf
        \EndFor
    \EndIf
\EndFor
\State \textbf{return} $G = (V, E)$
\end{algorithmic}
\end{algorithm}

%
%
\section{Methodology}\label{sec:3}

This section introduces the overall structure of the proposed RUL-aware motion planning framework,  
which extends the standard RRT* algorithm by incorporating joint health information into the planning and feedback process, forming a closed-loop architecture.

\subsection{Overview of the RUL-aware Motion Planning Framework}\label{sec:3.1}

\begin{figure}
    \centering
    \includegraphics[width=0.9\linewidth]{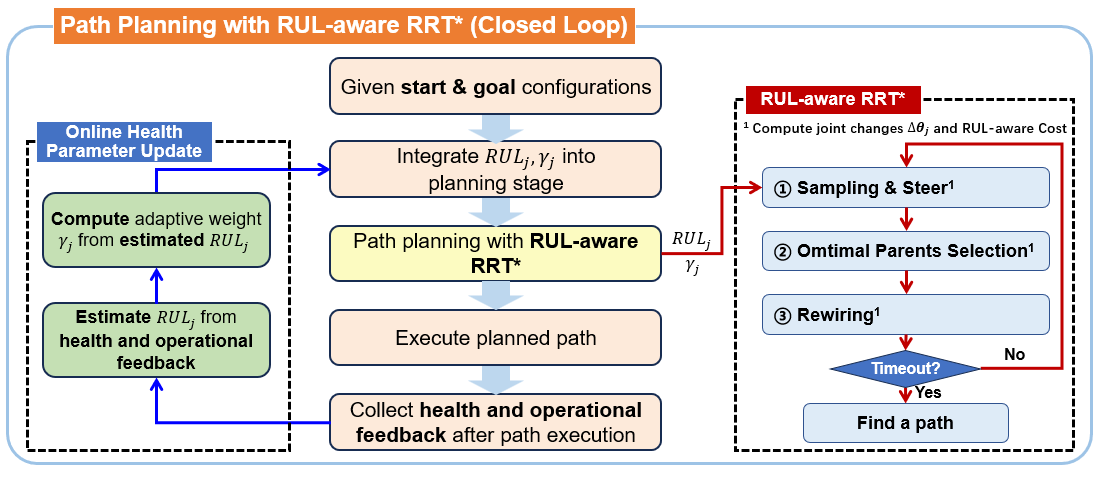}
    \caption{RUL-aware path planning with RRT* (closed-loop framework)}
    \label{fig:rulrrtstar}
\end{figure}

Figure~\ref{fig:rulrrtstar} provides an overview of the proposed closed-loop framework for path planning with RUL-aware RRT*. 
Instead of treating motion planning as a purely geometric problem, we embed the planner into a health-aware loop that repeatedly updates the joint Remaining Useful Life (RUL) and adapts the planning cost accordingly. 
For each task, the robot starts from a given pair of start and goal configurations and computes a collision-free path while explicitly considering the degradation state of each joint. The overall process establishes a closed-loop interaction between planning, execution, and health monitoring.

On the left, an online health-parameter module collects health and operational feedback after each path execution, such as joint usage, temperature or vibration measurements, and other condition-monitoring signals.  
From these data, the module estimates the joint \( \mathit{RUL_j} \) using any available degradation or prognostic model.  
A mapping function then converts the estimated \( \mathit{RUL_j} \) values into adaptive joint weights \( \gamma_j \), which quantify how much each joint should be penalized in the planning cost.  
This process can be further divided into two key mechanisms: the Online Joint RUL Update Mechanism and the RUL-driven Adaptive Joint Weighting Mechanism,  
which are responsible for the dynamic estimation of joint lifetime and the adaptive computation of health-based weighting, respectively.  
The detailed formulations of these mechanisms are presented in Sections~\ref{sec:3.3} and~\ref{sec:3.4}.

In the centre and on the right, the RUL-aware RRT* planner integrates \( \mathit{RUL_j} \) and \( \gamma_j \) into its cost function while preserving the standard RRT* pipeline of sampling, steering, optimal parent selection, rewiring, and timeout checking.  
The detailed definition of the RUL-aware cost formulation, which combines geometric efficiency with joint health awareness, is presented in Section~\ref{sec:3.2}.
During each iteration, the planner evaluates candidate edges using an RUL-aware cost that discourages excessive motion on highly degraded joints and promotes a more balanced distribution of joint-level motion usage, while still favouring short and feasible paths. After executing the planned path, new health and operational feedback are collected, and the RUL and weight parameters are updated for the next planning episode. 
Therefore, RUL and \( \gamma \) are applied in three stages: 
(i) pre-planning estimation and weighting, 
(ii) in-planning cost evaluation, and 
(iii) post-planning update from feedback, forming a closed loop between motion planning and health awareness.

\begin{algorithm}[t]
\caption{RUL-aware RRT* Motion Planning}
\label{alg:rulrrt}
\begin{algorithmic}[1]
\State $V \gets \{x_{\mathrm{init}}\}$; \quad $E \gets \emptyset$
\For{$i = 1,\dots,n$}
    \State $RUL \gets \textsc{UpdateRUL}()$ 
        \Comment{Online RUL update mechanism}
    \State $\gamma \gets \textsc{UpdateJointWeights}(RUL)$ 
        \Comment{RUL-driven adaptive joint weighting}
    \State $x_{\mathrm{rand}} \gets \textsc{SampleFree}()$
    \State $x_{\mathrm{nearest}} \gets \textsc{Nearest}(G=(V,E), x_{\mathrm{rand}}, RUL, \gamma)$
        \Comment{Stage~1: RUL/$\gamma$-weighted nearest neighbor selection}
    \State $x_{\mathrm{new}} \gets \textsc{Steer}(x_{\mathrm{nearest}}, x_{\mathrm{rand}})$
    \If{$\textsc{ObstacleFree}(x_{\mathrm{nearest}}, x_{\mathrm{new}})$}
        \State $X_{\mathrm{near}} \gets \textsc{Near}\big(G=(V,E), x_{\mathrm{new}},
        \min\{\gamma_{\mathrm{RRT^*}}(\log(\mathrm{card}(V))/\mathrm{card}(V))^{1/d}, \eta\}\big)$
        \State $V \gets V \cup \{x_{\mathrm{new}}\}$
        \State $x_{\min} \gets x_{\mathrm{nearest}}$
        \State $c_{\min} \gets 
            \textsc{Cost}(x_{\mathrm{nearest}};\, RUL, \gamma)
            + \textsc{Cost}(\text{Line}(x_{\mathrm{nearest}}, x_{\mathrm{new}});\,
            RUL, \gamma)$
        \Comment{Stage~2: parent selection under RUL-aware cost}
        \ForAll{$x_{\mathrm{near}} \in X_{\mathrm{near}}$}
            \If{$\textsc{CollisionFree}(x_{\mathrm{near}}, x_{\mathrm{new}})$
            $\wedge$
            $\textsc{Cost}(x_{\mathrm{near}};\, RUL, \gamma)
            + \textsc{Cost}(\text{Line}(x_{\mathrm{near}}, x_{\mathrm{new}});\,
            RUL, \gamma)
            < c_{\min}$}
                \State $x_{\min} \gets x_{\mathrm{near}}$
                \State $c_{\min} \gets 
                    \textsc{Cost}(x_{\mathrm{near}};\, RUL, \gamma)
                    + \textsc{Cost}(\text{Line}(x_{\mathrm{near}}, x_{\mathrm{new}});\,
                    RUL, \gamma)$
            \EndIf
        \EndFor
        \State $E \gets E \cup \{(x_{\min}, x_{\mathrm{new}})\}$
        \ForAll{$x_{\mathrm{near}} \in X_{\mathrm{near}}$}
            \If{$\textsc{CollisionFree}(x_{\mathrm{new}}, x_{\mathrm{near}})$
            $\wedge$
            $\textsc{Cost}(x_{\mathrm{new}};\, RUL, \gamma)
            + \textsc{Cost}(\text{Line}(x_{\mathrm{new}}, x_{\mathrm{near}});\,
            RUL, \gamma)
            < \textsc{Cost}(x_{\mathrm{near}};\, RUL, \gamma)$}
                \State $x_{\mathrm{parent}} \gets \textsc{Parent}(x_{\mathrm{near}})$
                \State $E \gets (E \setminus \{(x_{\mathrm{parent}}, x_{\mathrm{near}})\})
                \cup \{(x_{\mathrm{new}}, x_{\mathrm{near}})\}$
                \Comment{Stage~3: rewiring under RUL-aware cost}
            \EndIf
        \EndFor
    \EndIf
\EndFor
\State \Return $G = (V,E)$
\end{algorithmic}
\end{algorithm}

\subsection{RUL-aware Cost Formulation}\label{sec:3.2}

In the standard RRT* framework, the local cost between two configurations depends solely on the geometric distance in the joint space, where all joints are implicitly considered to have equal importance in motion generation.
Although this approach ensures geometric efficiency, it neglects the fact that different joints may be in different health conditions. 
As a result, a geometrically shortest path may repeatedly use joints that are already degraded, leading to faster wear and earlier system failure.

To address this limitation and enable health-aware decision-making during path generation, 
the proposed RUL-aware cost formulation integrates the Remaining Useful Life (RUL) of each joint into the cost evaluation process. 
The cost between two configurations \(x_i\) and \(x_{i+1}\) is defined as:
\begin{equation}
\mathrm{Cost}(x_i, x_{i+1}) =
\alpha \sum_{j=1}^{J} |\Delta q_j|
+
\lambda \sum_{j=1}^{J} 
\frac{\gamma_j |\Delta q_j|}{\max(\mathit{RUL_j}, R_{\min})},
\end{equation}

where \( \Delta q_j = q_{x_{i+1},j} - q_{x_i,j} \) represents the displacement of joint \( j \), 
and \( \alpha \) and \( \lambda \) are coefficients that balance geometric optimality and health awareness. 
The first term corresponds to the conventional geometric distance in joint space and maintains path efficiency when the robot is in healthy condition. 
The second term introduces a health-related penalty that increases the cost for joints with shorter remaining life. 
Here, \( \gamma_j \) is the adaptive weight derived from the RUL-driven weighting mechanism, 
\( \mathit{RUL_j} \) denotes the remaining useful life of joint \( j \), 
and \( R_{\min} \) is a small lower bound preventing division singularity.

Through this modification, the planner no longer selects paths solely based on geometric length. 
Instead, it also considers the current health state of each joint, giving preference to motions that distribute workload more evenly across the robot. 
In this way, the RUL-aware cost encourages the use of healthier joints, reduces the stress on degraded ones, 
and achieves a controllable balance between path efficiency and long-term reliability. 
This mechanism contributes to more balanced degradation among joints and delays the occurrence of the first joint failure.

\subsection{Online Joint RUL Update Mechanism}\label{sec:3.3}

The online joint RUL update mechanism plays a pivotal role in linking motion planning with health awareness. 
Rather than treating RUL estimation as an isolated prognostic task, 
this mechanism integrates the health-update process into the motion-planning loop, 
enabling the planner to continuously adapt to the evolving degradation of each joint. 
After every planning–execution cycle, the robot collects health and operational data (e.g., joint motion, load, temperature, vibration), which reflect the current wear state of each actuator. These feedback signals are used to update the joint-level remaining useful life (RUL), ensuring that the next planning iteration starts with the most recent health information. The updated RUL values are then converted into adaptive joint weights \( \gamma_j \), which directly influence the RUL-aware RRT* cost function and consequently reshape the motion decisions. Through this mechanism, motion planning becomes an iterative, health-driven process rather than a static optimization problem.

To express this process generically, the RUL update can be formulated as:
\begin{equation}
\mathit{RUL_j^{(t+1)}} = 
\Phi \big( \mathit{RUL_j^{(t)}}, \; 
\mathcal{H}_j^{(t)}, \; 
\mathcal{O}_j^{(t)} \big),
\label{eq:rul_update_general}
\end{equation}
where \( \Phi(\cdot) \) denotes a general update function integrating the historical RUL state \( \mathit{RUL_j^{(t)}} \), 
the health feedback \( \mathcal{H}_j^{(t)} \), and the operational feedback \( \mathcal{O}_j^{(t)} \), 
with \( \mathcal{H}_j^{(t)} \) and \( \mathcal{O}_j^{(t)} \) representing the health and operational features extracted from sensor feedback, respectively.
Here, \( t \) indexes each planning–execution episode rather than continuous time. 
This formulation supports different update strategies, including:

\begin{itemize}
    \item \textbf{Prognostics-based:} 
    \( \Phi(\cdot) = \mathcal{M}_{\text{pred}}(\mathcal{H}_j^{(t)}, \mathcal{O}_j^{(t)}) \), 
    where \( \mathcal{M}_{\text{pred}} \) is a predictive model estimating RUL directly from condition-monitoring data.
    \item \textbf{Degradation-based:} 
    \( \Phi(\cdot) = \mathit{RUL_j^{(t)}} - f(\Delta\theta_j^{(t)}) \), 
    where \( f(\cdot) \) quantifies degradation caused by incremental joint usage \( \Delta\theta_j^{(t)} \).
    \item \textbf{Hybrid recursive update:} 
    \( \Phi(\cdot) = (1-\alpha)\mathit{RUL_j^{(t)}} + \alpha\,\mathcal{M}_{\text{pred}}(\mathcal{H}_j^{(t)}, \mathcal{O}_j^{(t)}) \),
    where \( \alpha \in [0,1] \) balances model-based prediction and experience-based degradation.
\end{itemize}

This mechanism transforms RUL estimation from an offline diagnostic process into an online, 
task-level adaptation strategy that dynamically closes the loop between health monitoring and motion planning. 
By maintaining continuously updated health awareness, 
the proposed framework achieves self-adaptive planning behavior that promotes balanced degradation and extended system lifetime.

The resulting updated RUL values serve as the basis for the adaptive joint weighting mechanism discussed in Section~\ref{sec:3.4}.

\subsection{RUL-driven Adaptive Joint Weighting Mechanism}\label{sec:3.4}

Based on the updated joint RUL values obtained from the online update mechanism, an adaptive joint weighting process is applied to reflect the current health condition of each actuator in motion planning. 
In the standard RRT* formulation, all joints are treated equally in the cost computation, which overlooks the fact that different joints may degrade at different rates. 
To overcome this limitation, the proposed mechanism adjusts the relative weights of joints according to their remaining useful life (RUL), assigning higher weights to joints with lower RUL and lower weights to healthier ones. 
This allows the planner to distribute motion effort more reasonably among the joints while keeping the overall cost strength unchanged.

To quantify how uneven the joint health distribution is, we introduce an imbalance index \(B\) based on the coefficient of variation of the joint RUL values:
\begin{equation}
B = \frac{\mathrm{std}(\mathrm{RUL})}{\mathrm{mean}(\mathrm{RUL}) + \varepsilon},
\label{eq:B}
\end{equation}
where \(\mathrm{std}(\cdot)\) and \(\mathrm{mean}(\cdot)\) represent the standard deviation and mean of all joint RUL values, and \(\varepsilon\) is a small constant for numerical stability. 
The index \(B\) measures the dispersion of the RUL distribution: a larger \(B\) indicates a more pronounced difference between healthy and weak joints.

The imbalance index is then mapped to a sharpness parameter \(p\) that controls how aggressively the weighting focuses on weak joints:
\begin{equation}
p = 1 + B,
\label{eq:p}
\end{equation}
When the RUL values are similar (\(B \approx 0\)), the parameter \(p\) is close to 1 and the weighting difference between joints is small. 
As the health imbalance increases, \(p > 1\) produces a more peaked allocation that emphasizes joints with shorter RUL.

Given this parameter, the raw weights for each joint are defined as
\begin{equation}
w_j = \frac{1}{\big(\max(\mathit{RUL_j}, R_{\min})\big)^{p}},
\label{eq:w}
\end{equation}
where \(R_{\min}\) is a small lower limit to avoid numerical issues when RUL becomes very small. 
This definition ensures that joints with smaller RUL receive larger raw weights, with the contrast amplified as \(p\) increases. 
To keep the overall cost strength consistent, the raw weights are normalised as
\begin{equation}
\gamma_j = \frac{J\, w_j}{\sum_{i=1}^{J} w_i},
\label{eq:gamma}
\end{equation}
so that the relative distribution of weights changes according to the health condition, but the average weight remains equal to one. 
In this way, the influence of joint degradation on the cost is adjusted without affecting the overall magnitude of the planner’s objective.

When all joints have similar RUL values, the resulting weights \(\gamma_j\) are nearly uniform, and the planner behaves like a standard geometric planner. 
When one or more joints have much lower RUL, their corresponding \(\gamma_j\) increase, which makes movements involving those joints more costly. 
As a result, the planner tends to use healthier joints more often and reduces the workload on the weaker ones. 
This adaptive weighting mechanism helps the system maintain a better balance of joint usage and slows down the degradation of vulnerable components over multiple planning cycles, contributing to longer and more reliable operation.

\subsection{Integration with RRT}\label{sec:3.5}

The proposed RUL-aware mechanisms are integrated into the standard RRT* algorithm through modifications to its three key stages: Expansion, Parent Selection, and Rewiring.  
These changes allow the planner to consider both geometric efficiency and joint health during tree construction, while maintaining the sampling-based nature and asymptotic optimality of RRT*.

During the Expansion stage, the nearest node is no longer determined purely by Euclidean distance in the configuration space.  
Instead, the selection process incorporates the RUL-aware cost, enabling the planner to favor neighbors that lead to smoother joint utilization and lower cumulative degradation.  
This adjustment prevents the planner from repeatedly extending the tree through heavily degraded joints, thereby reducing stress concentration on weak actuators.

In the Parent Selection stage, the accumulated cost from the start configuration to each candidate parent node is evaluated using the health-aware cost formulation described earlier.  
This allows the planner to choose a parent not only based on geometric proximity but also on the health condition of the joints involved in the connecting path.  
Consequently, motion paths that impose less effort on degraded joints are naturally preferred.

During the Rewiring stage, the same health-aware cost is used to re-evaluate local connections in the tree. By checking whether the total cost of neighboring nodes can be reduced under the RUL-aware criterion, the planner incrementally refines the tree structure toward paths that achieve both motion efficiency and balanced joint usage. This ensures that even after the initial connection, local adjustments continue to optimize the distribution of joint-level motion usage as the tree grows.

After each planning cycle, the updated RUL values and joint weighting coefficients \(\gamma_j\) are fed back to the next iteration, forming a closed loop between motion planning and condition monitoring.  
Through this continuous feedback, the planner dynamically adapts to the robot’s evolving health condition, ensuring that motion trajectories remain both efficient and reliability-aware over time.

%
%
\section{Case Study}\label{sec:4}

This case study provides a comprehensive evaluation of the proposed RUL-aware RRT*, focusing on its effectiveness in mitigating degradation imbalance across multiple operational scenarios. Through controlled experiments, we quantify its impact on reliability, failure progression, and long-term task execution capability.

\subsection{Experimental Setup}\label{sec:4.1}

The proposed RUL-aware motion planning framework is implemented and evaluated within a simulation environment to ensure controllability and repeatability of the experiments. 
All implementations are carried out in the ROS Melodic framework using the MoveIt motion planning package integrated with the Open Motion Planning Library (OMPL). 
A 6-DOF UR5 robotic manipulator is used as the test platform, providing sufficient kinematic complexity to evaluate joint-level health adaptation. 
The proposed planner is embedded in OMPL as a customized RRT* variant that incorporates the RUL-aware cost formulation and adaptive joint weighting mechanisms introduced in Sections~\ref{sec:3.2}–\ref{sec:3.4}. 
All planning and execution processes are visualized and simulated in RViz, which offers collision checking, trajectory interpolation, and real-time visualization of the robot’s joint motion.

\begin{figure}
    \centering
    \includegraphics[width=0.4\linewidth]{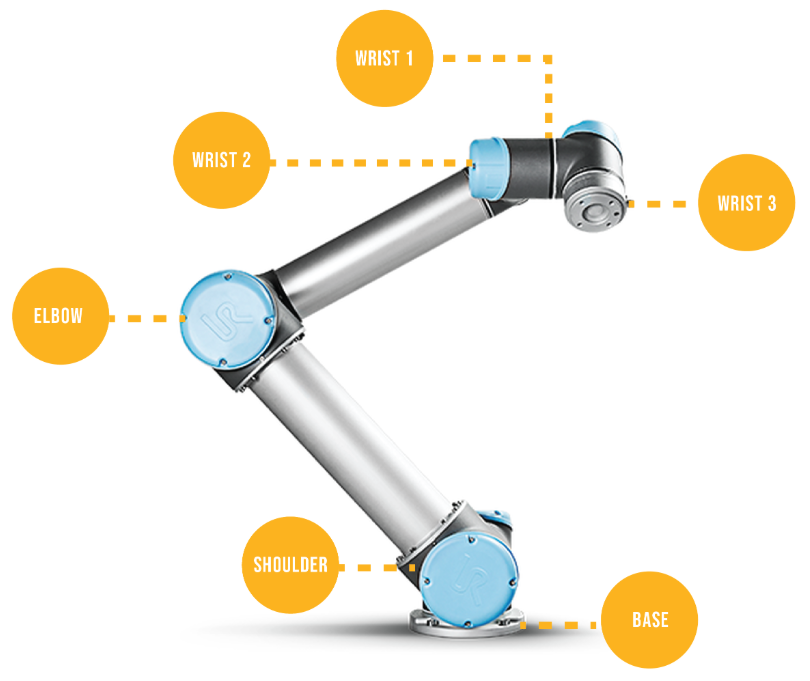}
    \caption{Experimental 6-DOF robotic arm used for validation}
    \label{fig:placeholder}
\end{figure}

Since the experiments are performed entirely in simulation, no real sensor measurements such as torque, temperature, or vibration are collected. 
Instead, the health state of each joint is abstracted through its angular displacement, which reflects the cumulative mechanical usage over repeated operations. 
After each pick-and-place task, the incremental joint angle changes are recorded and used as feedback to update the estimated Remaining Useful Life (RUL) of each joint through the online update mechanism. 
This simulation-based feedback design allows controlled testing of the planner’s adaptive behaviour without introducing uncertainty from real sensor noise. 
In this way, the angular usage serves as a surrogate for physical degradation, enabling the evaluation of how the planner dynamically responds to health variations across multiple planning and execution cycles.

Both the baseline RRT* and the proposed RUL-aware RRT* share identical sampling strategies, step sizes, and connection radii to ensure fair comparison. 
The weighting coefficients in the cost formulation are set to \(\alpha = 1\) and \(\lambda = 1\), 
providing a balanced trade-off between geometric optimality and health awareness. 
Each experiment operates in a closed-loop manner, where the planner generates a path, the simulated robot executes the motion, 
and the updated RUL and weighting parameters are fed back for the next planning iteration. 
This iterative process forms a self-contained evaluation loop that mimics long-term operation under progressive degradation while maintaining full control of experimental conditions. 
The specific degradation configurations and test scenarios used for validation are presented in Section~\ref{sec:4.2}.

\subsection{Test Scenarios}\label{sec:4.2}

\begin{table}
\centering
\caption{Summary of test scenarios for RUL-aware motion planning experiments.}
\label{tab:test_scenarios}
\begin{tabular}{p{3cm} p{4.5cm} p{2cm} p{5cm}}
\hline
\textbf{Scenario} & \textbf{Initial RUL Configuration} & \textbf{No. of Tasks} & \textbf{Evaluation Objective} \\
\hline
\textit{Full Health Scenario} & [1000, 1000, 1000, 1000, 1000, 1000] & 80 & Verify whether the planner can maintain balanced joint utilization when all joints start equally healthy, preventing early overuse and achieving synchronized degradation. \\[3pt]

\textit{Heterogeneous Degradation Scenario} & [500, 800, 400, 600, 1000, 700] & 80 & Evaluate the planner’s ability to redistribute joint-level motion usage under non-uniform initial RULs, ensuring adaptive usage balancing and delayed degradation. \\[3pt]

\textit{Local Degradation Scenario} & [100, 1000, 1000, 1000, 1000, 1000] & Until first failure & Test whether the planner can protect the severely degraded joint, reduce its workload, and extend the system’s first failure time while maintaining task feasibility. \\
\hline
\end{tabular}
\end{table}

To comprehensively evaluate the performance of the proposed RUL-aware motion planning framework, 
three groups of experiments, as listed in Table~\ref{tab:test_scenarios}, 
are designed to simulate different joint degradation conditions and verify the planner’s adaptability under varying health states. 
All experiments are conducted on the same UR5 manipulator introduced in Section~\ref{sec:4.1}, 
and each task follows a pick-and-place motion, where the start and goal configurations are randomly generated within the robot’s reachable workspace. 
Each scenario represents a distinct reliability condition, covering uniform aging, uneven wear, and localized failure.

According to the general RUL update framework introduced in Section~\ref{sec:2}, 
the remaining lifetime of each joint evolves iteratively based on its health and operational feedback. 
In this experimental study, the update function $\Phi(\cdot)$ is implemented in a \textit{degradation-based} form, 
where the RUL evolution depends explicitly on the cumulative joint usage $\theta_j$ \citep{heng2009rotating, si2011remaining}. 
This approach corresponds to the degradation-based branch of the general formulation and is represented by the following closed-form model:

\begin{equation}
RUL_j^{(t)} = R_{\min} + (R_0 - R_{\min})
\left(1 - \frac{\theta_j^{(t)}}{\theta_{\max}}\right)^{p}.
\label{eq:rul_degradation}
\end{equation}

In this model, \(R_0 = 1000\) represents the initial remaining life of a healthy joint, 
and \(R_{\min} = 0\) defines the minimum bound to prevent negative values.  
The parameter \( \theta_{\max} = 1000 \) specifies the maximum cumulative angular usage corresponding to complete lifetime consumption.  
After each pick-and-place task \(t\), the accumulated joint displacement is updated as
\begin{equation}
\theta_j^{(t+1)} = \theta_j^{(t)} + \Delta\theta_j^{(t)}
\end{equation}
where \( \Delta\theta_j^{(t)} \) is the actual joint rotation measured during task \(t\).  
Substituting this value into Eq.~\eqref{eq:rul_degradation} yields the new \(RUL_j^{(t+1)}\), 
realizing the degradation-based RUL update process within the general framework.  
This formulation establishes a direct link between mechanical usage and lifetime consumption, 
enabling the planner to adapt motion allocation according to the real-time health condition of each joint.

The exponent \(p\) in Eq.~\eqref{eq:rul_degradation} controls the curvature of the degradation trajectory 
and allows different wear characteristics to be emulated.  
When \(p = 1\), degradation proceeds linearly with usage, representing uniform wear.  
When \(p = 1.5\), degradation accelerates during the late stage, simulating components that wear rapidly near failure.  
When \(p = 0.8\), degradation is sublinear, indicating slow early wear and gradual lifetime decay.  
These parameter settings enable flexible simulation of diverse physical aging patterns.

\begin{figure}
    \centering
    \includegraphics[width=0.6\linewidth]{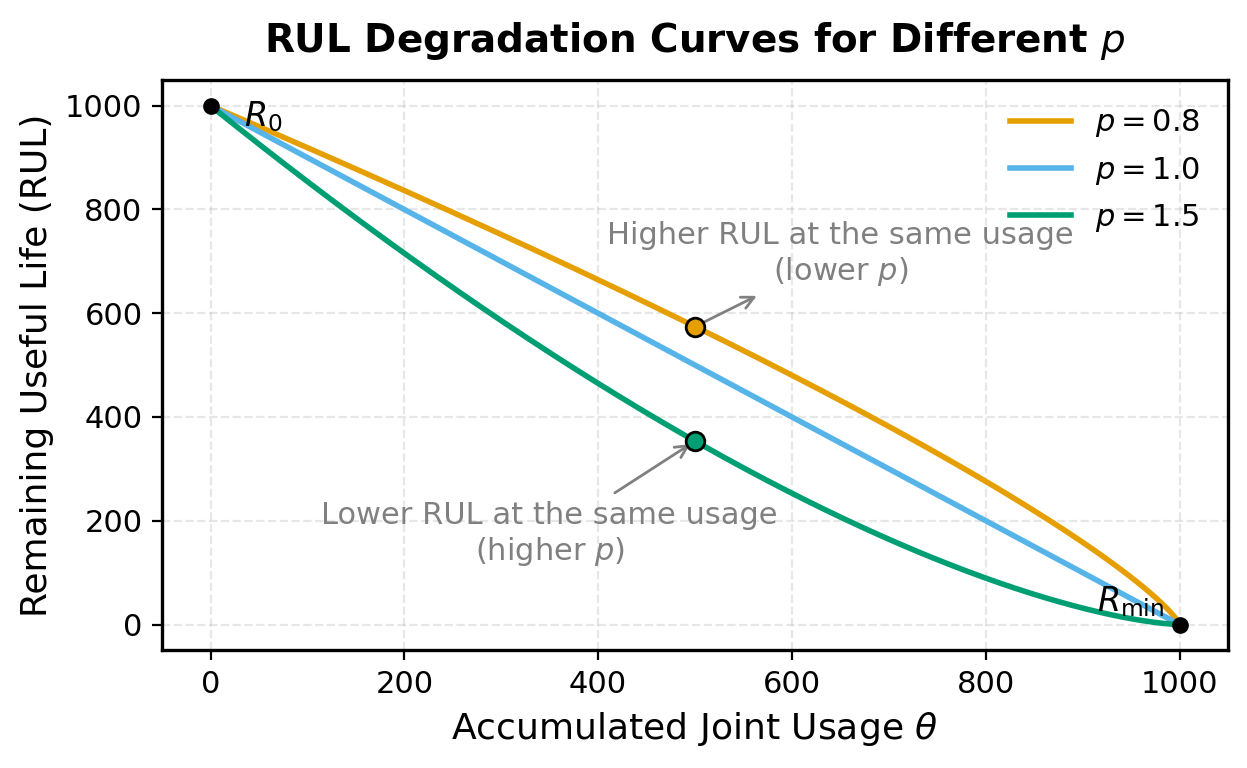}
    \caption{RUL degradation curves for different exponent values \(p\). 
    A smaller \(p\) produces slower early degradation and a more gradual lifetime decline, 
    while a larger \(p\) leads to accelerated wear near the end of life.}
    \label{fig:rul_curves}
\end{figure}

In all cases, when \( \theta_j = \theta_{\max} \), the corresponding \( RUL_j = R_{\min} \) marks the failure threshold.  
The experiments are executed either for 80 consecutive pick-and-place tasks or until the first joint reaches its minimum RUL value (\( RUL_j \le R_{\mathrm{fail}} = 0 \)), 
representing system-level failure.

The first scenario, \textit{Full Health Scenario}, initializes all six joints with equal lifetimes of [1000, 1000, 1000, 1000, 1000, 1000].  
This configuration reflects a robot operating under fully healthy conditions.  
It aims to test whether the proposed planner can maintain balanced joint utilization and prevent premature overuse of any specific joint.  
Ideally, all joints should degrade synchronously, producing a uniform decline of RUL over time.

The second scenario, \textit{Heterogeneous Degradation Scenario}, sets the initial RUL configuration to [500, 800, 400, 600, 1000, 700], representing a manipulator where some joints have experienced prior wear.  
This case examines the planner’s ability to redistribute motion workloads according to the current health diversity, 
assigning lighter loads to degraded joints and heavier ones to healthy joints in order to achieve lifetime balancing and delayed deterioration.

The third scenario, \textit{Local Degradation Scenario}, introduces a severely weakened actuator with an initial RUL of \( [100, 1000, 1000, 1000, 1000, 1000] \).  
This extreme case assesses the planner’s capability to protect the weakest joint, limit its further usage, and extend the system’s overall operational time.  
Unlike the other two settings, this simulation continues until the weakest joint reaches failure (\( RUL_j = 0 \)), 
thereby quantifying how effectively the RUL-aware strategy delays the onset of the first component failure.

Collectively, these three test scenarios enable a comprehensive evaluation of the proposed RUL-aware planner under various degradation conditions.  
They capture the full spectrum of realistic health states, including uniform aging, uneven degradation, and localized failure.  
This experimental design provides a basis for quantitative assessment of joint-level motion redistribution, degradation balancing, and lifetime extension effects. 
The evaluation metrics used to analyze these scenarios are presented in Section~\ref{sec:4.3}.

\subsection{Evaluation Metrics}\label{sec:4.3}

To systematically evaluate the proposed RUL-aware motion planning framework, 
a unified set of lifetime- and balance-oriented indicators is established. 
All metrics are derived from the joint-level remaining useful life \(RUL_j^{(t)}\) 
and the accumulated angular displacement \(\theta_j^{(t)}\), 
where \(j \in \{1, \dots, 6\}\) indexes the robot joints and \(t\) denotes the task index. 
These indicators collectively quantify the planner’s ability to extend system lifetime, 
maintain balanced degradation, and achieve efficient workload distribution.

\vspace{4pt}
\noindent\textbf{(1) Total RUL Consumption (\(\Delta RUL_{\mathrm{sum}}\)).}  
\begin{equation}
\Delta RUL_{\mathrm{sum}} = \sum_{j=1}^{6} \big(RUL_j^{(0)} - RUL_j^{(T)}\big)
\end{equation}
This metric measures the total amount of lifetime consumed by all joints after executing \(T\) tasks.  
Here, \(RUL_j^{(0)}\) and \(RUL_j^{(T)}\) denote the remaining useful life of joint \(j\) 
at the beginning and at the end of the experiment, respectively.  
A smaller \(\Delta RUL_{\mathrm{sum}}\) indicates lower cumulative lifetime consumption, 
reflecting improved lifetime efficiency.

\vspace{4pt}
\noindent\textbf{(2) Remaining RUL Sum (\(RUL_{\mathrm{sum}}^{(T)}\)).}  
\begin{equation}
RUL_{\mathrm{sum}}^{(T)} = \sum_{j=1}^{6} RUL_j^{(T)}
\end{equation}
This indicator quantifies the total residual lifetime of the system after task \(T\).  
A higher \(RUL_{\mathrm{sum}}^{(T)}\) suggests that the system remains healthier 
after performing the same number of tasks.

\vspace{4pt}
\noindent\textbf{(3) RUL Degradation Balance (\(CV_{\mathrm{RUL}}^{(T)}\)).}  
\begin{equation}
CV_{\mathrm{RUL}}^{(T)} = 
\frac{\sigma(RUL_1^{(T)}, \dots, RUL_6^{(T)})}
     {\mu(RUL_1^{(T)}, \dots, RUL_6^{(T)})}
\end{equation}
This coefficient of variation measures how uniformly the joints degrade 
in terms of remaining useful life.  
\(\mu(\cdot)\) and \(\sigma(\cdot)\) denote the mean and standard deviation, respectively.  
A smaller \(CV_{\mathrm{RUL}}^{(T)}\) indicates more synchronized degradation 
and better degradation balance across the six joints.

\vspace{4pt}
\noindent\textbf{(4) System Lifetime Bottleneck (\(RUL_{\min}^{(T)}\)).}  
\begin{equation}
RUL_{\min}^{(T)} = \min_j RUL_j^{(T)}
\end{equation}
This metric represents the remaining lifetime of the most degraded (weakest) joint.  
Since the system lifetime is limited by its earliest failing component, 
a larger \(RUL_{\min}^{(T)}\) reflects an extended system-level reliability.

\vspace{4pt}
\noindent\textbf{(5) System Lifetime Recovery Ratio (\(G_{\mathrm{rec}}\)).}  
\begin{equation}
G_{\mathrm{rec}} = 
\frac{RUL_{\min}^{(T),\mathrm{RUL}} - RUL_{\min}^{(T),\mathrm{base}}}
     {RUL_{\min}^{(T),\mathrm{base}}} \times 100\%
\end{equation}
This metric evaluates how much the RUL-aware planner improves 
the weakest joint lifetime compared with the baseline planner 
after completing the same number of tasks.  
\(RUL_{\min}^{(T),\mathrm{RUL}}\) and \(RUL_{\min}^{(T),\mathrm{base}}\) 
are the minimum joint RUL values at task \(T\) under the RUL-aware and baseline planners, respectively.  
A higher \(G_{\mathrm{rec}}\) indicates stronger lifetime recovery capability.

\vspace{4pt}
\noindent\textbf{(6) First Failure Task Index (\(k_{\mathrm{fail}}\)).}  
\begin{equation}
k_{\mathrm{fail}} = \min\{t \mid RUL_{\min}^{(t)} \le R_{\mathrm{fail}}\}
\end{equation}
This indicator denotes the number of tasks completed before any joint 
reaches the failure threshold \(R_{\mathrm{fail}}\).  
A larger \(k_{\mathrm{fail}}\) means the system can perform more tasks 
before its first failure occurs.

\vspace{4pt}
\noindent\textbf{(7) Relative Lifetime Gain (\(G_{\mathrm{life}}\)).}  
\begin{equation}
G_{\mathrm{life}} = 
\frac{k_{\mathrm{fail}}^{\mathrm{RUL}} - k_{\mathrm{fail}}^{\mathrm{base}}}
     {k_{\mathrm{fail}}^{\mathrm{base}}} \times 100\%
\end{equation}
This metric compares how much longer the system lasts 
under the RUL-aware planner relative to the baseline in run-to-failure conditions.  
A higher \(G_{\mathrm{life}}\) represents a greater extension in system lifetime.

\vspace{4pt}
\noindent\textbf{(8) RUL Balance at Failure (\(CV_{\mathrm{RUL}}^{\mathrm{fail}}\)).}  
\begin{equation}
CV_{\mathrm{RUL}}^{\mathrm{fail}} =
\frac{\sigma(RUL_1^{(k_{\mathrm{fail}})}, \dots, RUL_6^{(k_{\mathrm{fail}})})}
     {\mu(RUL_1^{(k_{\mathrm{fail}})}, \dots, RUL_6^{(k_{\mathrm{fail}})})}
\end{equation}
This coefficient of variation quantifies the degradation uniformity among joints 
at the moment of system failure.  
A smaller value indicates that even when failure occurs, the degradation remains balanced.

\begin{table}
\centering
\caption{Summary of evaluation metrics and their applicability across different test scenarios. 
Applicable scenarios correspond to those defined in Section~\ref{sec:4.2}: 
\textit{Full Health Scenario}, \textit{Local Degradation Scenario}, and \textit{Heterogeneous Degradation Scenario}.}
\label{tab:metrics_summary}
\renewcommand{\arraystretch}{1.1}
\begin{tabularx}{\linewidth}{P{3.2cm} P{2.4cm} P{3.8cm} P{2.2cm} Y}
\hline
\textbf{Metric} & \textbf{Symbol} & \textbf{Physical Meaning} & \textbf{Tendency} & \textbf{Applicable Scenario(s)} \\
\hline
Total RUL Consumption & \(\Delta RUL_{\mathrm{sum}}\) & Total lifetime consumed by all joints & Lower is better & Full Health Scenario; Heterogeneous Degradation Scenario \\
Remaining RUL Sum & \(RUL_{\mathrm{sum}}^{(T)}\) & Total remaining lifetime after all tasks & Higher is better & Full Health Scenario; Heterogeneous Degradation Scenario \\
RUL Degradation Balance & \(CV_{\mathrm{RUL}}^{(T)}\) & Uniformity of degradation among joints & Lower is better & Full Health Scenario; Heterogeneous Degradation Scenario \\
System Lifetime Bottleneck & \(RUL_{\min}^{(T)}\) & Remaining lifetime of the weakest joint & Higher is better & Full Health Scenario; Heterogeneous Degradation Scenario \\
System Lifetime Recovery Ratio & \(G_{\mathrm{rec}}\) & Improvement of bottleneck lifetime vs. baseline & Higher is better & Full Health Scenario; Heterogeneous Degradation Scenario \\
First Failure Task Index & \(k_{\mathrm{fail}}\) & Number of tasks before the first joint failure & Higher is better & Local Degradation Scenario \\
Relative Lifetime Gain & \(G_{\mathrm{life}}\) & Lifetime extension compared with baseline & Higher is better & Local Degradation Scenario \\
RUL Balance at Failure & \(CV_{\mathrm{RUL}}^{\mathrm{fail}}\) & Uniformity of RUL at the failure moment & Lower is better & Local Degradation Scenario \\
\hline
\end{tabularx}
\end{table}

The physical meaning, desired tendency, and scenario applicability 
of all indicators are summarized in Table~\ref{tab:metrics_summary}. Overall, these metrics collectively provide a comprehensive evaluation of 
the RUL-aware planner’s performance across different reliability conditions: 
(i) prolonging lifetime and delaying the first failure under local degradation, 
(ii) maintaining synchronized and efficient degradation under full health, 
and (iii) adaptively balancing heterogeneous degradation to recover system reliability.

\subsection{Results and Discussion}\label{sec:4.4}

The experimental results across the three degradation scenarios collectively demonstrate the effectiveness of the proposed RUL-aware motion planning framework. The following discussion synthesizes the key findings and explains how the proposed mechanisms contribute to balanced degradation, adaptive workload redistribution, and lifetime extension.

\vspace{4pt}
\noindent\textbf{(1) Full Health Scenario}

\begin{figure}
    \centering
    \includegraphics[width=1\linewidth]{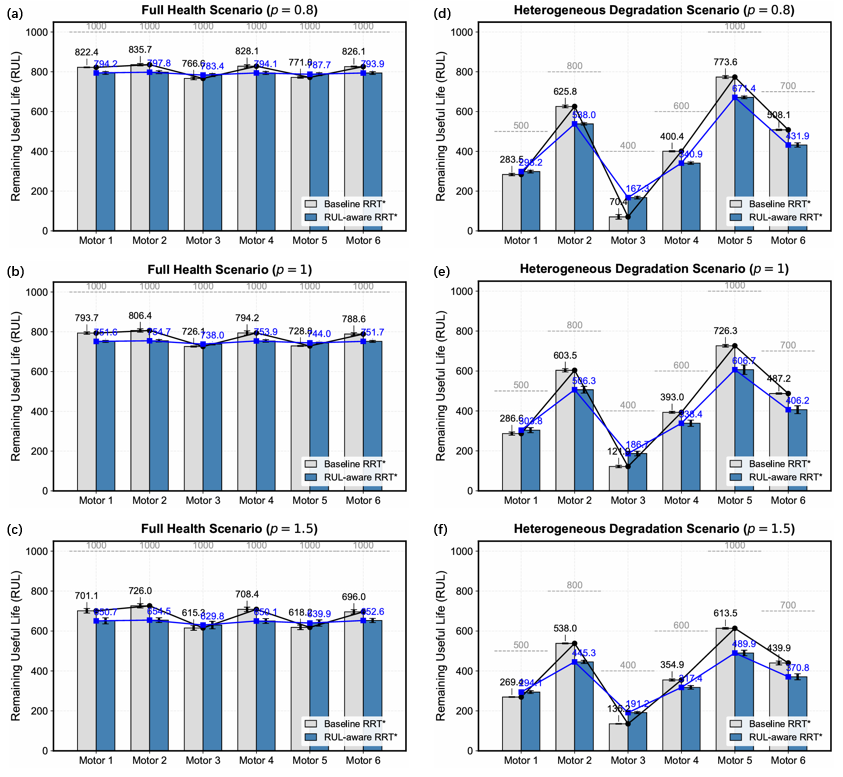}
    \caption{
        Comparison of the final remaining useful life (RUL) of six motors 
        under two system conditions: 
        \textbf{(a–c)}~Full Health Scenario (FHS) and 
        \textbf{(d–f)}~Heterogeneous Degradation Scenario (HDS), 
        for three parameter settings ($p = 0.8$, $1$, and $1.5$). 
        Each subplot compares the \textit{Baseline RRT*} and 
        \textit{RUL-aware RRT*} results for the same configuration.  
        Gray and blue bars represent the mean RUL values of the Baseline RRT* and 
        RUL-aware RRT* planners, respectively, with error bars indicating the standard deviation across runs.  
        The black and blue lines connect the mean RUL values of the Baseline and RUL-aware planners, respectively, 
        while the gray dashed horizontal lines denote the initial reference RUL for each motor.
    }
    \label{fig:HDS_HPS_RUL}
\end{figure}

Under the Full Health Scenario (FHS), all six joints start with identical initial lifetimes, allowing us to examine whether different planning strategies introduce degradation imbalance even when the system is fully symmetric. The final remaining useful life (RUL) distributions in Fig.~\ref{fig:HDS_HPS_RUL}(\(a\)--\(c\)) show that the Baseline RRT* produces substantial variation across joints for all degradation parameters \(p = 0.8, 1.0,\) and \(1.5\). For instance, at \(p = 0.8\), the weakest joint ends with a minimum remaining lifetime of \(RUL_{\min} \approx 704\), while the healthiest joint retains around 835, yielding a disparity of roughly 130 units. By contrast, the RUL-aware RRT* maintains the six joints within a much narrower band (approximately 730--790), with a significantly higher minimum remaining lifetime of \(RUL_{\min} \approx 733\). This improvement corresponds to a system lifetime recovery ratio of \(G_{\mathrm{rec}} \approx (733 - 704)/704 \approx 4.1\%\), and the corresponding recovery ratios for \(p = 1.0\) and \(p = 1.5\) are approximately \(1.7\%\) and \(0.5\%\), respectively, based on their respective increases in \(RUL_{\min}\). Although these percentages appear modest, the higher bottleneck values clearly indicate that the proposed planner provides meaningful protection against premature degradation concentration, effectively delaying the formation of the first component-level lifetime limit even when all joints begin equally healthy.

\begin{figure}
    \centering
    \includegraphics[width=1\linewidth]{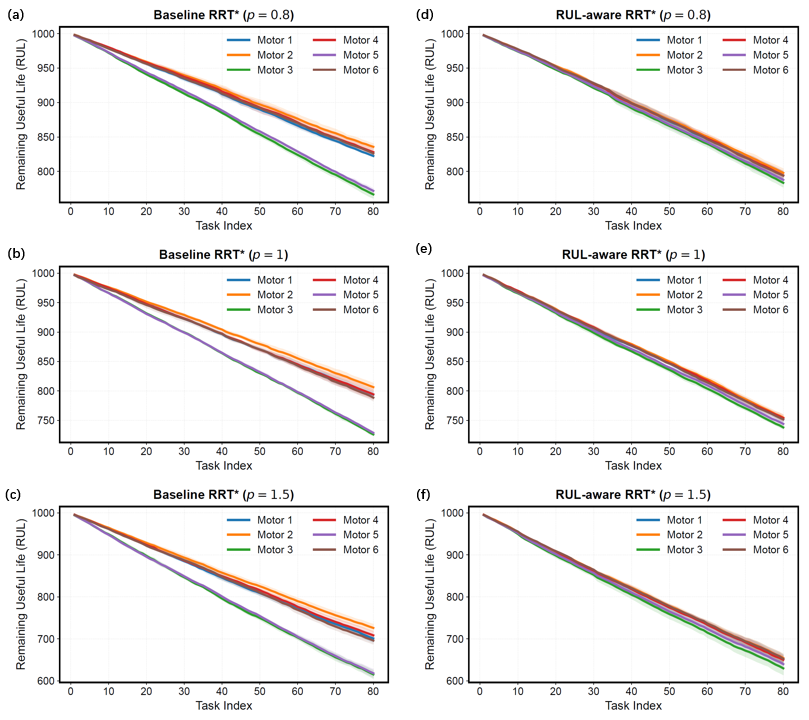}
    \caption{
        Evolution of the remaining useful life (RUL) of six motors under the 
        \textbf{Full Health Scenario (FHS)} condition across three parameter settings 
        ($p = 0.8$, $1$, and $1.5$). 
        Subfigures (a–c) show the baseline RRT* results, while (d–f) present the 
        RUL-aware RRT* outcomes under the same task sequences. 
        The shaded regions represent the standard deviation across multiple experimental runs.
        Compared with the baseline, the RUL-aware planner achieves more balanced RUL trajectories among all motors, indicating more balanced joint-level motion usage and slower degradation trends.
    }
    \label{fig:FHS_RUL_trajectory}
\end{figure}

These final-state differences are the cumulative consequence of how degradation evolves during task execution. As depicted in Fig.~\ref{fig:FHS_RUL_trajectory}, the RUL trajectories of the Baseline RRT* progressively diverge as tasks proceed, with different joints exhibiting different rates of decline. Without health-aware intervention, these discrepancies accumulate monotonically and ultimately produce the imbalances observed in the final RUL distributions. In contrast, the trajectories of the RUL-aware RRT* remain nearly overlapped across all \(p\) values, indicating that the closed-loop weighting mechanism continuously compensates for small usage differences as they emerge. Whenever a joint begins to accumulate slightly more motion, its weight increases, making it less favorable in subsequent planning steps and preventing deviation from amplifying over time.

\begin{figure}
    \centering
    \includegraphics[width=0.7\linewidth]{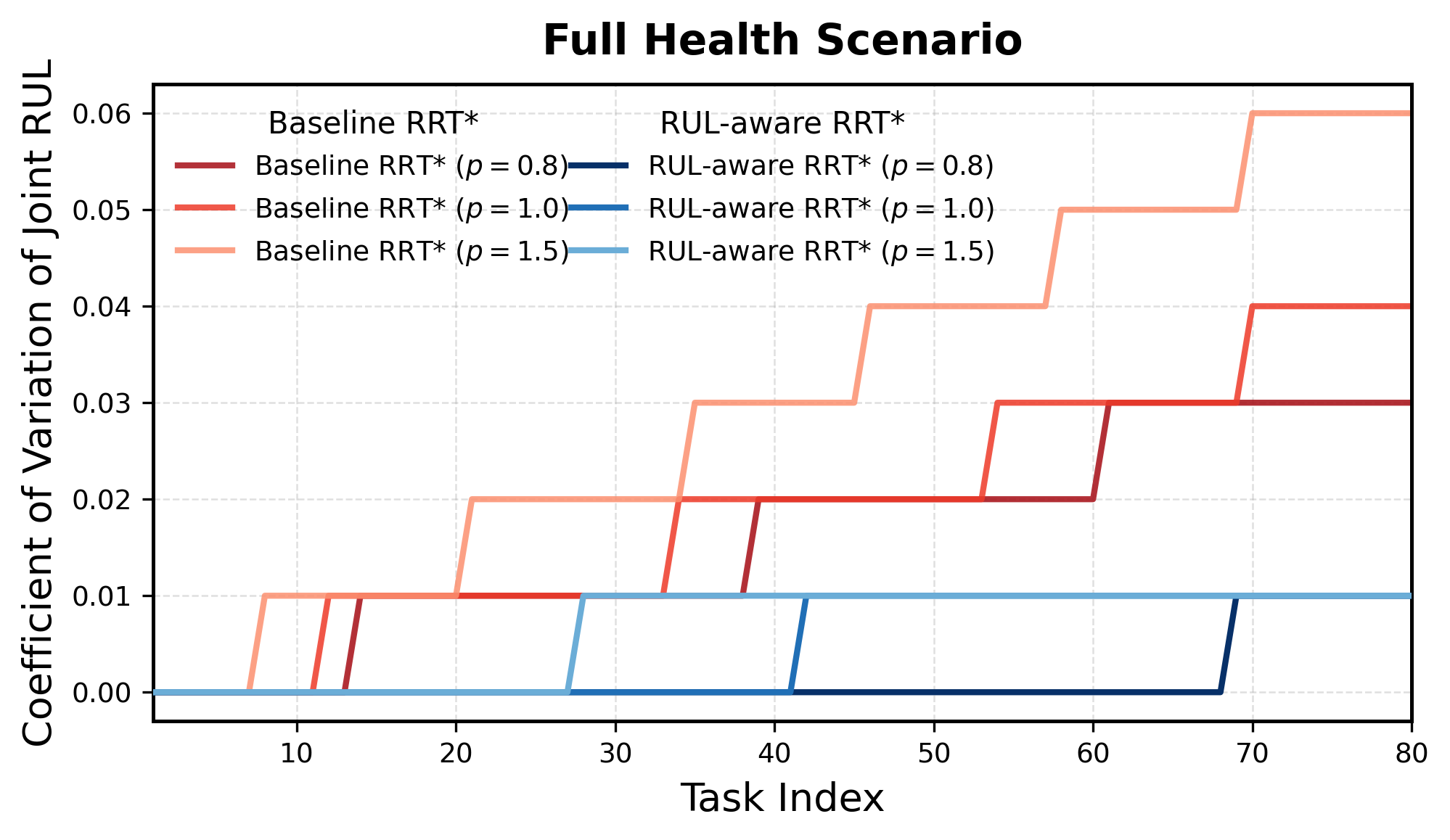}
    \caption{
        Evolution of the coefficient of variation (CV) of joint remaining useful life (RUL)
        in the \textbf{Full Health Scenario (FHS)} over task execution.  
        Each curve corresponds to a specific parameter setting ($p = 0.8$, $1$, or $1.5$), 
        comparing the \textit{Baseline RRT*} (red tones) and 
        \textit{RUL-aware RRT*} (blue tones) planners.  
        Shaded areas around each curve indicate the standard deviation across repeated runs.
    }
    \label{fig:FHS_cv}
\end{figure}

The evolution of degradation balance is quantitatively characterized in Fig.~\ref{fig:FHS_cv}, which plots the coefficient of variation of joint RUL throughout task execution. For the Baseline RRT*, the resulting \(CV_{\mathrm{RUL}}\) exhibits a staircase-like growth pattern, eventually reaching values between 0.03 and 0.06 depending on \(p\), indicating a progressively widening spread across joints. In contrast, the \(CV_{\mathrm{RUL}}\) under the RUL-aware planner stays near zero for most of the execution horizon and increases only marginally towards the end. This quantitative behavior aligns with the trajectory patterns in Fig.~\ref{fig:FHS_cv}: imbalance naturally emerges and amplifies under the baseline, whereas the RUL-aware planner stabilizes the relative degradation of all joints, effectively delaying the onset of a system lifetime bottleneck.

\begin{figure}
    \centering
    \includegraphics[width=0.7\linewidth]{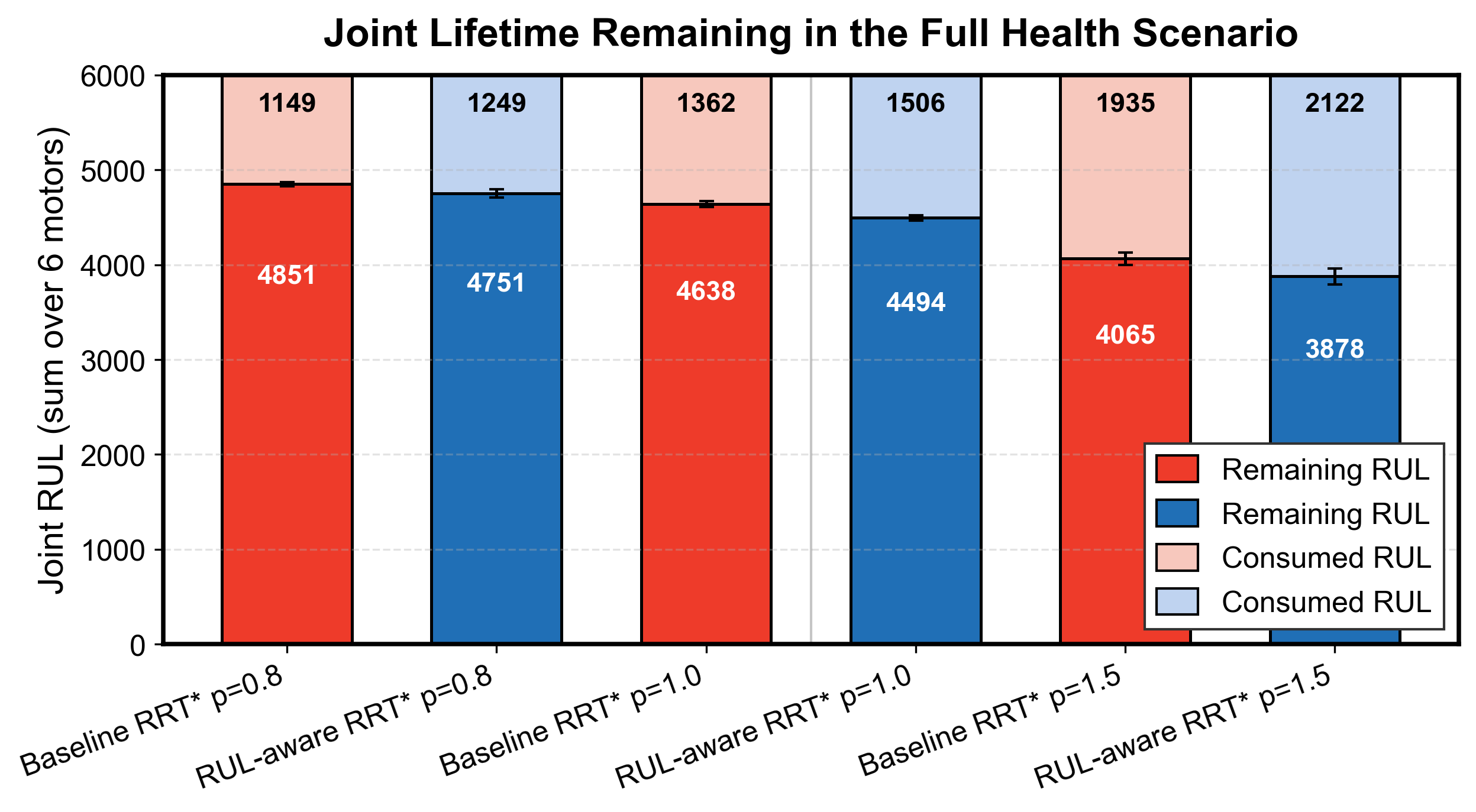}
    \caption{
        Remaining joint lifetime at the end of the \textbf{Full Health Scenario}. 
        For each degradation shape parameter ($p = 0.8$, $1.0$, and $1.5$), 
        the two adjacent bars correspond to the \textit{Baseline RRT*} 
        and the \textit{RUL-aware RRT*} planners. 
        Each bar is stacked: the bottom segment represents the sum of the remaining 
        RUL over the six joints, while the top segment shows the corresponding 
        consumed lifetime, with the total bar height normalized to $6000$. 
        Error bars on the remaining segment indicate the standard deviation 
        across repeated runs.
    }
    \label{fig:FHS_rul_sum_grouped}
\end{figure}

While the RUL-aware framework shifts a small amount of additional motion to healthier joints, its overall lifetime consumption remains modest. Fig.~\ref{fig:FHS_rul_sum_grouped} shows the stacked consumed and remaining lifetimes after 80 tasks. The total RUL consumption \(\Delta RUL_{\mathrm{sum}}\) increases slightly for the RUL-aware RRT*, typically by about 100--200 RUL units, but this minor increase yields substantial benefits in remaining lifetime. The final remaining RUL sum \(RUL_{\mathrm{sum}}^{(T)}\) improves by approximately 100, 238, and 287 units for \(p = 0.8, 1.0,\) and \(1.5\), respectively. In other words, the system ends in a uniformly healthier state despite having consumed marginally more lifetime, representing a favorable and interpretable trade-off between minimal additional wear and clearly improved reliability.

Overall, the results under the Full Health Scenario demonstrate that the RUL-aware RRT* proactively maintains balanced degradation, suppresses the formation of early bottlenecks, stabilizes degradation trajectories over long execution horizons, and yields higher remaining lifetime with only marginally increased consumption. These improvements arise not from isolated planning decisions but from the sustained effect of closed-loop health feedback, highlighting the robustness and generality of the proposed framework even when no joint is initially disadvantaged.

\vspace{4pt}
\noindent\textbf{(2) Heterogeneous Degradation Scenario}

Under the Heterogeneous Degradation Scenario (HDS), the six joints start with substantially different initial lifetimes, creating a setting in which degradation would naturally concentrate on the already weakened components if no health-aware mechanism is applied. The final RUL distributions in Fig.~\ref{fig:HDS_HPS_RUL}(\(d\)--\(f\)) clearly illustrate this effect for the Baseline RRT*: joints beginning with lower initial RUL degrade much faster and end with extremely small remaining lifetimes. For instance, at \(p = 0.8\), the weakest joint reaches only \(RUL_{\min} \approx 164\) after 80 tasks, while several healthier joints retain more than 600 units of RUL, yielding a disparity exceeding 450 units. In contrast, the RUL-aware RRT* significantly elevates the bottleneck lifetime to approximately \(RUL_{\min} \approx 203\), narrowing the spread across joints and reducing the severity of degradation concentration. This improvement corresponds to a system lifetime recovery ratio of \(G_{\mathrm{rec}} \approx (203 - 164)/164 \approx 23.8\%\), while the recovery ratios for \(p = 1.0\) and \(p = 1.5\) are approximately \(25.7\%\) and \(29.2\%\), respectively. These differences highlight the ability of the proposed planner to protect initially degraded joints and alleviate the system's lifetime bottleneck.

The divergence patterns visible in Fig.~\ref{fig:HDS_HPS_RUL} arise from the unequal motion loads accumulated during execution. In the Baseline RRT*, joints that begin with low health continue to shoulder significant portions of the motion, causing their degradation to accelerate and amplifying the initial heterogeneity. By contrast, the RUL-aware RRT* dynamically shifts effort toward healthier joints over the task sequence, reducing the degradation rate of weaker components. This health-driven redistribution prevents weak joints from collapsing prematurely and manifests directly in the elevated \(RUL_{\min}\) values observed at the end of execution.

\begin{figure}
    \centering
    \includegraphics[width=0.7\linewidth]{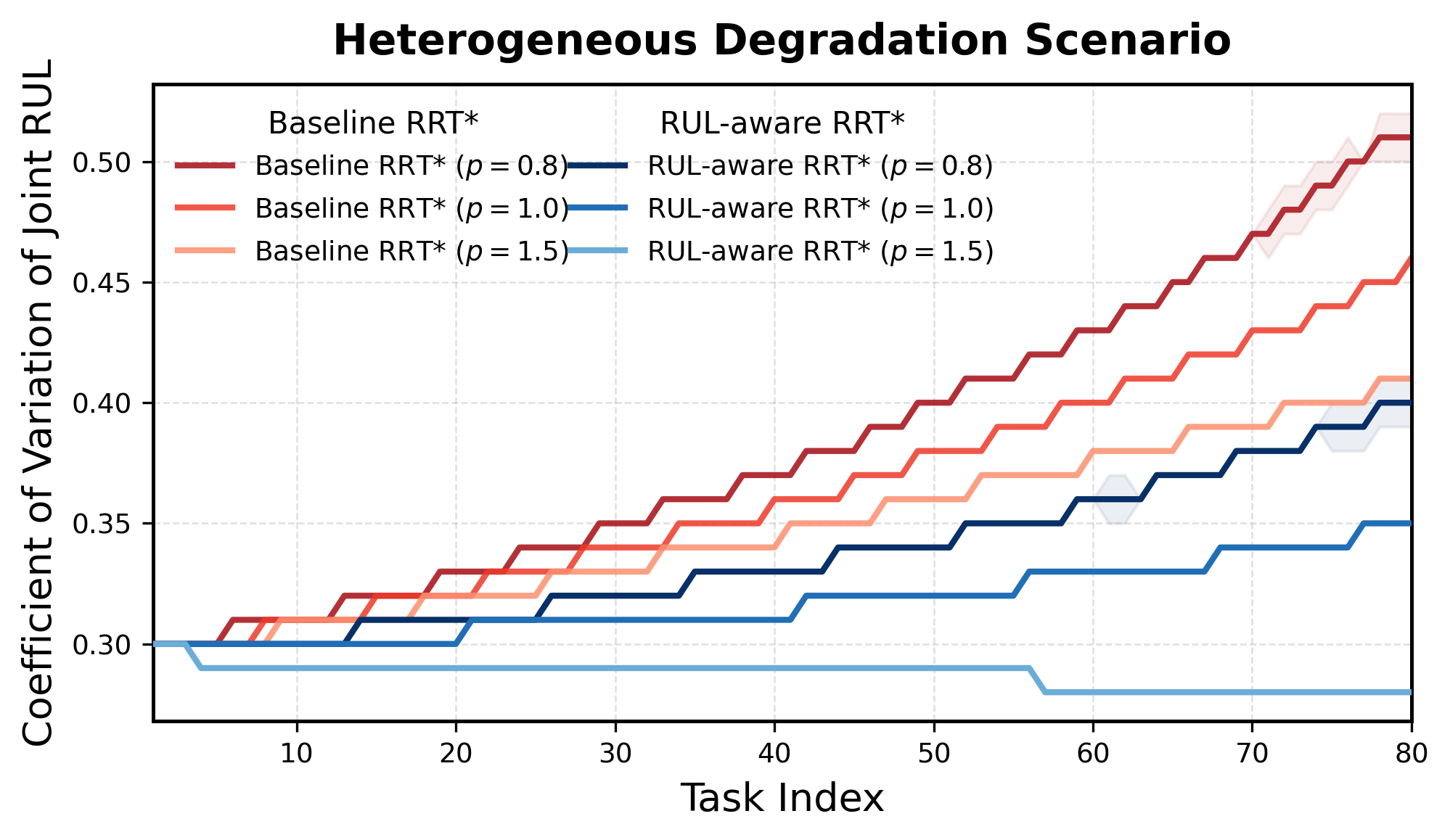}
    \caption{
        Evolution of the coefficient of variation (CV) of joint remaining useful life (RUL) 
        in the \textbf{Heterogeneous Degradation Scenario (HDS)} over task execution.  
        Each curve corresponds to a specific parameter setting ($p = 0.8$, $1$, or $1.5$), 
        comparing the \textit{Baseline RRT*} (red tones) and 
        \textit{RUL-aware RRT*} (blue tones) planners.  
        Shaded areas around each curve indicate the standard deviation across repeated runs.
    }
    \label{fig:HDS_cv}
\end{figure}

The quantitative evolution of degradation imbalance is further reflected in Fig.~\ref{fig:HDS_cv}, which reports the coefficient of variation of joint RUL across the task horizon. For the Baseline RRT*, the resulting \(CV_{\mathrm{RUL}}\) increases rapidly and reaches values in the range of 0.40--0.55 depending on \(p\), showing that the initial heterogeneity not only persists but intensifies over time. In contrast, the RUL-aware planner maintains substantially lower \(CV_{\mathrm{RUL}}\) values, typically within 0.25--0.35, demonstrating its ability to suppress the amplification of degradation imbalance. This behavior is consistent with the protective effect observed in the final RUL distributions and confirms that the proposed method mitigates the tendency for weak joints to deteriorate disproportionately.

\begin{figure}
    \centering
    \includegraphics[width=0.7\linewidth]{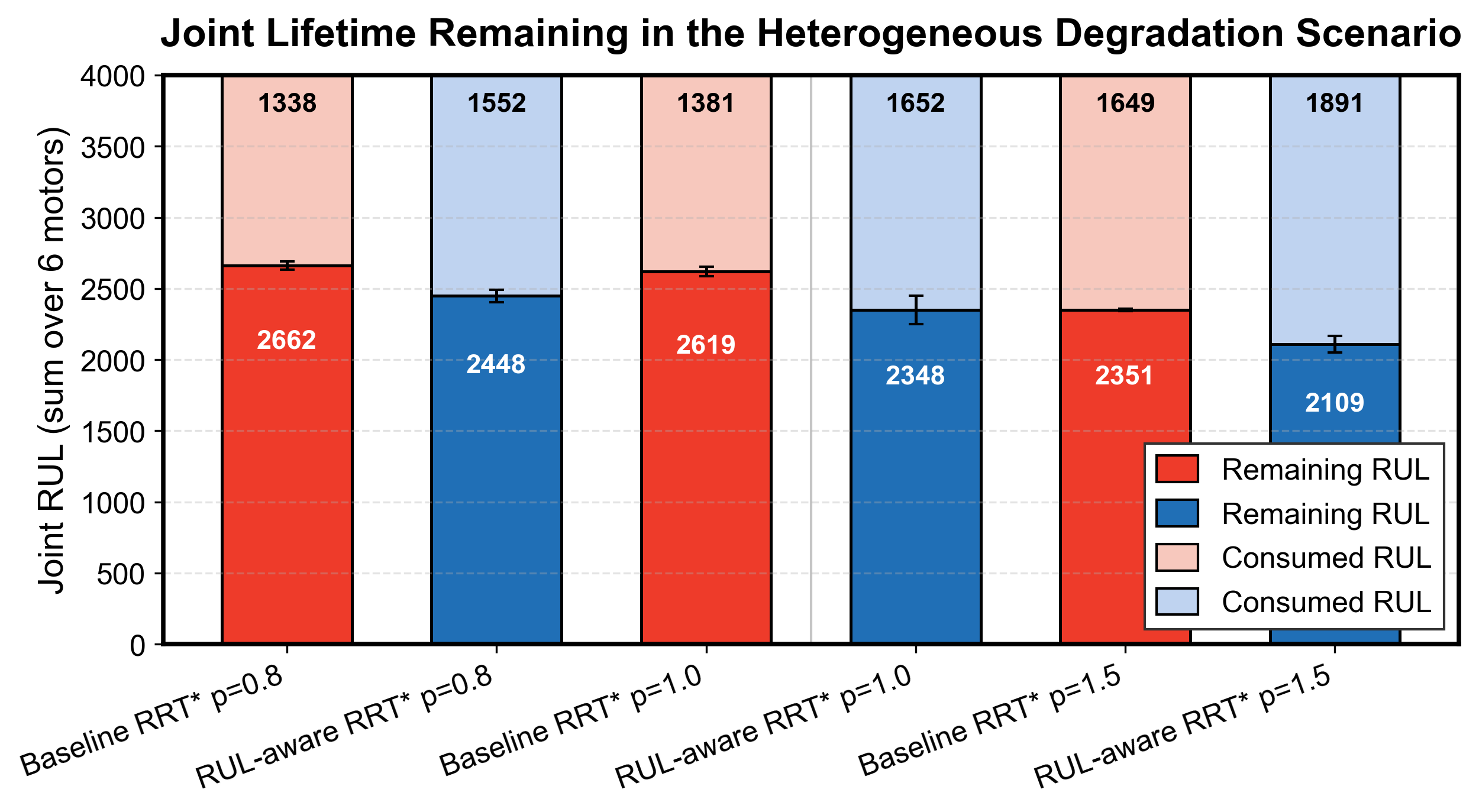}
    \caption{
        Remaining joint lifetime at the end of the \textbf{Heterogeneous Degradation Scenario}. 
        For each degradation shape parameter ($p = 0.8$, $1.0$, and $1.5$), 
        the two adjacent bars correspond to the \textit{Baseline RRT*} 
        and the \textit{RUL-aware RRT*} planners. 
        Each bar is stacked: the bottom segment represents the sum of the remaining 
        RUL across the six joints, while the top segment shows the consumed portion, 
        with the total bar height normalized to the initial heterogeneous lifetime 
        ($500 + 800 + 400 + 600 + 1000 + 700 = 4000$). 
        Error bars on the remaining segment indicate the standard deviation 
        across repeated runs.
    }
    \label{fig:HDS_rul_sum_grouped}
\end{figure}

Although the RUL-aware planner redistributes motion and increases the workload of healthier joints, its overall lifetime consumption remains modest relative to the gains achieved. As shown in Fig.~\ref{fig:HDS_rul_sum_grouped}, the total RUL consumption \(\Delta RUL_{\mathrm{sum}}\) is only slightly higher for the RUL-aware RRT*, typically by 80--150 RUL units. However, the remaining lifetime sum \(RUL_{\mathrm{sum}}^{(T)}\) improves significantly, increasing by approximately 185, 273, and 302 units for \(p = 0.8, 1.0,\) and \(1.5\), respectively. This reflects a favorable trade-off: a small increase in system-wide lifetime usage results in a much healthier system state and a substantially higher lifetime margin for the bottleneck component.

Overall, the results under the Heterogeneous Degradation Scenario demonstrate that the RUL-aware RRT* effectively counteracts the natural tendency of degradation to concentrate on initially weak joints. By redistributing motion according to joint health, the method raises the lifetime of the system bottleneck, slows the decline of degraded components, moderates the growth of degradation imbalance, and improves the system’s overall remaining lifetime. These outcomes arise from continuous closed-loop health feedback and highlight the planner’s strong capability to maintain system reliability even when the initial health distribution is highly uneven.

\vspace{4pt}
\noindent\textbf{(3) Local Degradation Scenario}

\begin{figure}
    \centering
    \includegraphics[width=0.7\linewidth]{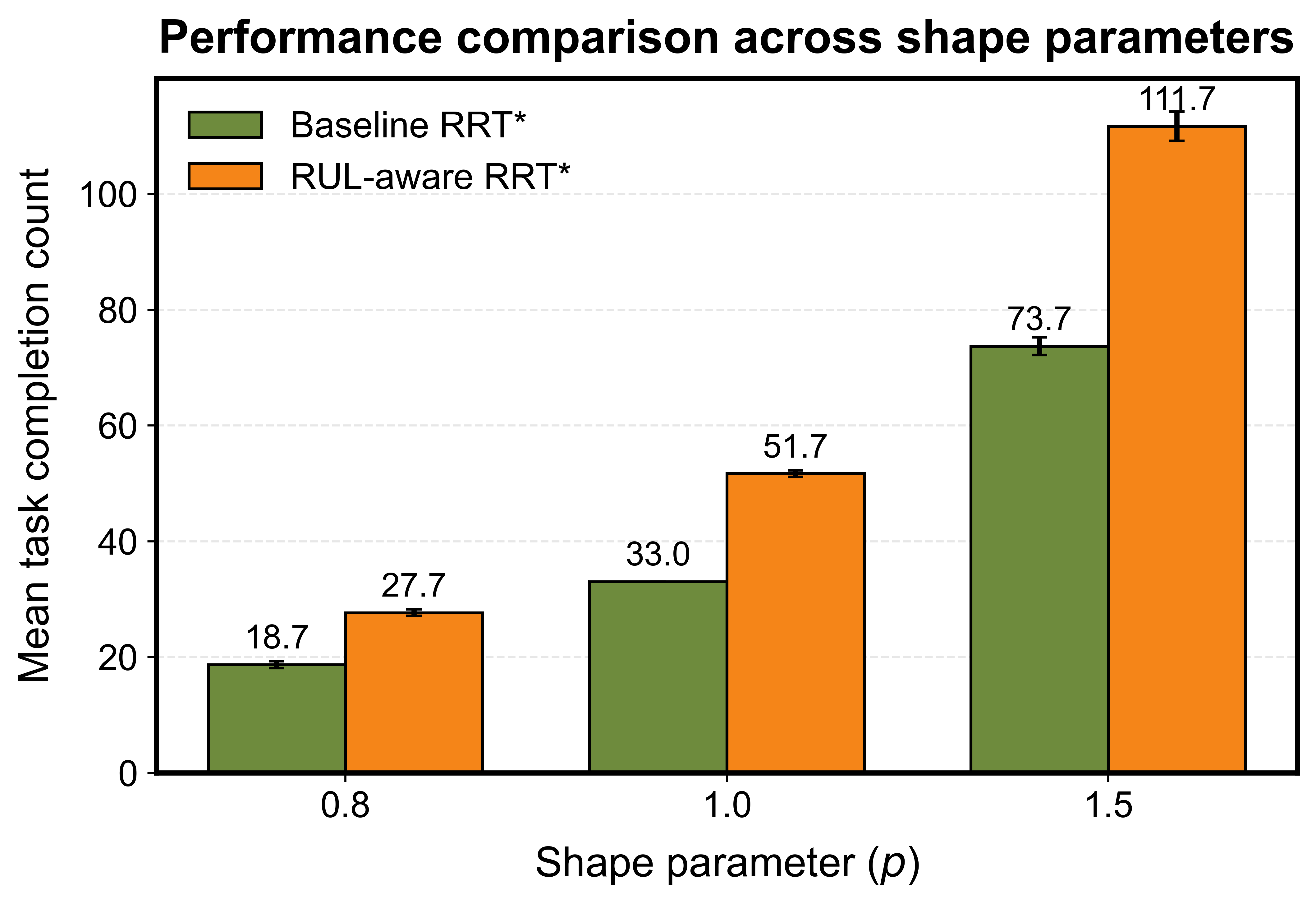}
    \caption{
        Comparison of the number of completed tasks for the 
        \textit{Baseline RRT*} and \textit{RUL-aware RRT*} planners 
        under the \textbf{Local Degradation Scenario (LDS)}.  
        Each group of bars corresponds to a different shape parameter ($p = 0.8$, $1$, and $1.5$).  
        Gray and blue bars represent the mean task completion counts of the Baseline and 
        RUL-aware planners, respectively, with black error bars showing the standard deviation 
        across repeated runs.  Numeric values above the bars indicate the mean values.
    }
    \label{fig:LDS_task_completion}
\end{figure}

Under the Local Degradation Scenario (LDS), the system begins with a clearly weakened joint whose initial remaining lifetime is significantly lower than that of the others, causing the overall system lifetime to be dominated by this bottleneck component. This scenario therefore directly evaluates whether a planning strategy can delay the occurrence of the first joint failure and extend task execution under run-to-failure conditions. As shown in Fig.~\ref{fig:LDS_task_completion}, the number of tasks completed before failure differs markedly between the two planners across all three degradation parameters \(p = 0.8, 1.0,\) and \(1.5\). For the Baseline RRT\*, the absence of any health-awareness mechanism results in the weak joint carrying substantial motion early in the execution, causing its RUL to drop rapidly. For instance, at \(p = 0.8\), the baseline completes only about \(18.7\) tasks before failure, whereas the RUL-aware RRT\* achieves approximately \(27.7\) tasks, corresponding to a relative lifetime gain of \(G_{\mathrm{life}} \approx (27.7 - 18.7)/18.7 \approx 48.1\%\). At \(p = 1.0\), the baseline completes about \(33.0\) tasks, while the RUL-aware RRT\* reaches approximately \(51.7\) tasks, giving a relative lifetime gain of \(G_{\mathrm{life}} \approx (51.7 - 33.0)/33.0 \approx 56.7\%\). At \(p = 1.5\), the baseline completes \(73.7\) tasks compared to \(111.7\) tasks under the RUL-aware planner, yielding \(G_{\mathrm{life}} \approx (111.7 - 73.7)/73.7 \approx 51.6\%\). This indicates that the proposed planner consistently increases the first failure task index \(k_{\mathrm{fail}}\) regardless of the degradation curve shape.

\begin{figure}
    \centering
    \includegraphics[width=0.7\linewidth]{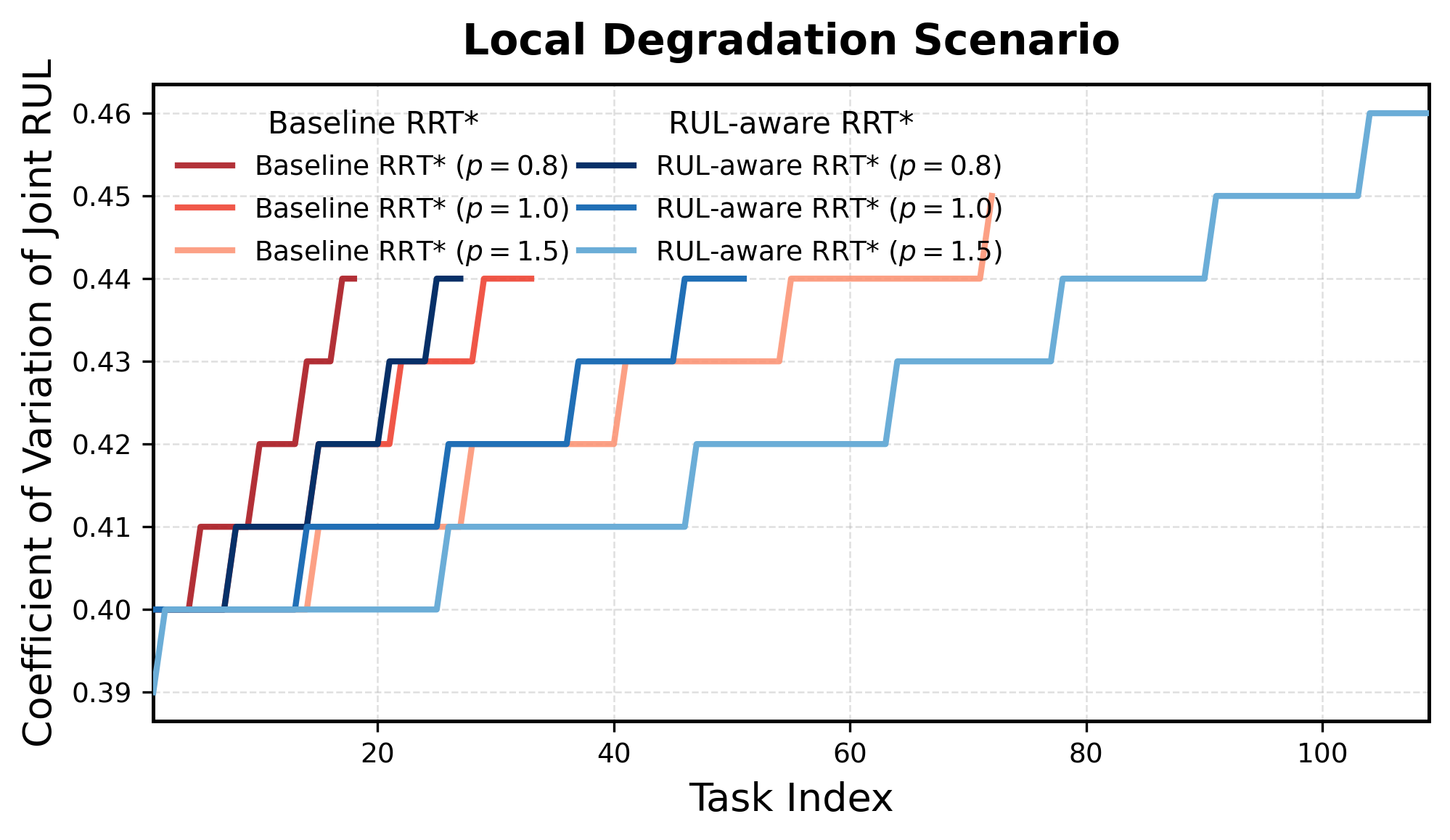}
    \caption{
        Evolution of the coefficient of variation (CV) of joint remaining useful life (RUL)
        in the \textbf{Local Degradation Scenario (LDS)} over task execution.  
        Each curve corresponds to a specific parameter setting ($p = 0.8$, $1$, or $1.5$), 
        comparing the \textit{Baseline RRT*} (red tones) and 
        \textit{RUL-aware RRT*} (blue tones) planners.  
        Shaded areas around each curve indicate the standard deviation across repeated runs.
    }
    \label{fig:LDS_cv}
\end{figure}

These improvements stem from fundamentally different degradation behaviors of the weakened joint. In the baseline planner, the weak joint continues to receive considerable use throughout the task sequence, quickly pushing it to the failure threshold. In contrast, the RUL-aware RRT* continuously increases the joint’s weight as its health declines, reducing its involvement in subsequent trajectories and significantly slowing its degradation rate. This effect is clearly reflected in Fig.~\ref{fig:LDS_cv}, where the Baseline RRT* shows a rapid rise in degradation imbalance: its \(CV_{\mathrm{RUL}}\) curve increases sharply early on and eventually reaches approximately 0.44–0.46 near failure, indicating that the weak joint suffers disproportionately accelerated degradation. The RUL-aware RRT*, by comparison, maintains much lower imbalance levels, with \(CV_{\mathrm{RUL}}\) remaining around 0.40–0.42, demonstrating that degradation differences are effectively suppressed throughout the execution horizon.

The superiority of the proposed method becomes even more evident when considering the degradation uniformity at the failure moment. The metric \(CV_{\mathrm{RUL}}^{\mathrm{fail}}\) characterizes the spread of joint RUL values at the instant the system reaches failure; a smaller value indicates a more balanced degradation state even at the point of collapse. Based on the trends observed in Fig.~\ref{fig:LDS_cv}, the RUL-aware RRT* achieves noticeably lower \(CV_{\mathrm{RUL}}^{\mathrm{fail}}\) compared with the baseline, confirming that the proposed method not only delays the failure of the weak joint (higher \(k_{\mathrm{fail}}\)) but also ensures a more uniform degradation distribution at failure, thus exhibiting stronger reliability margins.

Overall, the results under the Local Degradation Scenario demonstrate that the RUL-aware RRT* substantially prolongs system operation, increases the first failure task index \(k_{\mathrm{fail}}\), provides significant lifetime extension quantified by \(G_{\mathrm{life}}\), and maintains more balanced degradation at failure as reflected in the reduced \(CV_{\mathrm{RUL}}^{\mathrm{fail}}\). These advantages arise from the continuous closed-loop health feedback mechanism, which actively protects the initially weakened joint under run-to-failure conditions, thereby providing notable improvements in task completion capability and overall system reliability.

Across all three scenarios, the RUL-aware planner consistently outperforms the baseline, but the magnitude of improvement is not uniform across the degradation shape parameters \(p\). This is expected, because \(p\) controls how sensitively the RUL responds to cumulative usage in the degradation law, and thus how much “leverage’’ a health-aware planner can exert. When degradation is close to linear or relatively mild (smaller \(p\)), additional motion assigned to a stressed joint causes only moderate extra damage, so redistributing joint-level motion usage yields noticeable but limited gains, as reflected by the modest recovery ratios in the FHS and the moderate lifetime gains in the LDS. As \(p\) increases, the degradation law becomes more nonlinear: extra usage on an already weak joint incurs disproportionately large RUL loss. In this regime, the Baseline RRT* suffers severely because bottleneck joints collapse quickly in both HDS and LDS, whereas the RUL-aware planner leverages its reweighting mechanism to shield these joints. As a result, the proposed method yields substantially larger improvements in \(RUL_{\min}\), \(G_{\mathrm{rec}}\), and \(G_{\mathrm{life}}\) for intermediate and higher \(p\) values. The overall pattern suggests that the proposed method is robust across different degradation regimes, yet becomes particularly beneficial when the underlying physics is strongly damage-accelerating, i.e., when “using the wrong joint’’ is especially costly. This provides a practical insight for predictive maintenance: health-aware planning is most critical in systems where component degradation is highly nonlinear with respect to accumulated motion usage, yet it still provides noticeable benefits even under milder wear dynamics.

%
%
\section{Conclusion and Future Work}\label{sec:5}

This study examined the effectiveness of the RUL-aware RRT*, a motion-planning method that integrates joint-level remaining useful life (RUL) information into the planning process and adaptively adjusts joint usage based on evolving health conditions. Comprehensive evaluations under three representative scenarios, namely the Full Health Scenario (FHS), the Heterogeneous Degradation Scenario (HDS), and the Local Degradation Scenario (LDS), show that the RUL-aware RRT* consistently improves system reliability compared with the baseline RRT*.

In the Full Health Scenario, where all joints begin with identical lifetimes, the RUL-aware RRT* maintains balanced degradation throughout execution. It achieves higher minimum remaining lifetime \(RUL_{\min}\), smaller variation in joint RUL, and improved system lifetime recovery ratios \(G_{\mathrm{rec}}\), while requiring only a small increase in total lifetime consumption.

In the Heterogeneous Degradation Scenario, the benefits are more substantial. The method effectively protects initially weak joints and reduces the amplification of health disparities. It produces higher bottleneck lifetimes, more uniform degradation distributions, and lower \(CV_{\mathrm{RUL}}\), resulting in a healthier system state at the end of execution.

In the Local Degradation Scenario, which represents run-to-failure conditions, the RUL-aware RRT* significantly extends the system’s operational duration. By limiting the motion assigned to the degraded joint, it increases the first failure task index \(k_{\mathrm{fail}}\) by 48–56\% across different degradation parameters and achieves meaningful relative lifetime gains \(G_{\mathrm{life}}\). The degradation balance at failure is also improved, as reflected by smaller values of \(CV_{\mathrm{RUL}}^{\mathrm{fail}}\).

Although the RUL-aware RRT* consistently improves reliability, the magnitude of improvement varies with the degradation shape parameter \(p\). The gains become larger when degradation is more sensitive to accumulated motion usage, indicating that health-aware planning is particularly important in systems where small increases in joint usage lead to disproportionately large degradation. Even in milder degradation regimes, the method still provides noticeable benefits.

Overall, the results demonstrate that the RUL-aware RRT* effectively balances degradation, delays the formation of bottlenecks, and improves the lifetime performance of robotic systems operating under wear. By incorporating health feedback directly into the planning process, the method provides a practical and robust approach for enhancing long-term system reliability.

As a direction for future work, we plan to extend the RUL-aware RRT* to multi-robot settings in which several manipulators collaborate on shared tasks. Such systems introduce additional challenges in task allocation, motion coordination, and cross-robot health interaction. To address these complexities, we aim to incorporate agentic AI and reinforcement learning to enable adaptive motion allocation, coordinated planning, and dynamic health-aware decision making across multiple robotic arms. This direction has the potential to further enhance long-term system reliability in complex, multi-agent operational environments.

%
%
\section{Declaration of generative AI and AI-assisted technologies in the writing process}
During the preparation of this work the author(s) used ChatGPT 4o to partially refine the completed manuscript. After using this tool, the author(s) reviewed and edited the content as needed and take(s) full responsibility for the content of the published article.

%
%
\section{Declaration of Competing Interest}  
The authors declare that they have no known competing financial interests or personal relationships that could have appeared to influence the work reported in this paper.

%
%
\section{Acknowledgements}
Haibo Li acknowledges the support of Program of China Scholarship Council (202307090014). The research of Zhiguo Zeng is partically funded by the French Research Council under contract number ANR-22-CE-10-0004 (Project DFT).

\bibliographystyle{plainnat}

\bibliography{cas-refs}

\end{document}